\documentclass{article}

\usepackage[UKenglish]{babel}
\usepackage[numbers]{natbib}
\usepackage[nonatbib,final]{neurips_2020}

\usepackage[utf8]{inputenc} 
\usepackage[T1]{fontenc}    
\usepackage{hyperref}       
\usepackage{url}            
\usepackage{booktabs}       
\usepackage{amsfonts}       
\usepackage{nicefrac}       
\usepackage{microtype}      
\usepackage[table]{xcolor}
\usepackage{wrapfig}
\usepackage{amsmath}
\usepackage{amssymb}
\usepackage{float}
\usepackage{tcolorbox}
\usepackage{multirow}
\usepackage[capitalise]{cleveref}
\usepackage{pifont} 
\usepackage{changepage}  
\usepackage{pdflscape} 
\usepackage{caption}
\usepackage{subcaption}
\usepackage{stackengine} 
\usepackage{makecell}

\DeclareMathOperator*{\expect}{\mathbb E}
\newcommand\bmname{MAGICAL}
\newcommand\bc{{\rm bc}}
\newcommand\adv{{\rm adv}}
\newcommand\mt{{\rm mt}}
\newcommand\ST[1]{{\color{red}\textbf{ST:} {#1}}}

\newcommand\crule[1]{\textcolor[rgb]{#1}{\rule{0.2cm}{0.3cm}}}

\newcommand\getcurrentref[1]{%
 \ifnumequal{\value{#1}}{0}
  {??}
  {\the\value{#1}}%
}
\definecolor{darkblue}{rgb}{0.42,0.53,0.64}
\definecolor{lightblue}{rgb}{0.89,0.93,0.98}
\definecolor{boxframe}{rgb}{0.89,0.71,0.64}
\definecolor{boxfill}{rgb}{0.96,0.91,0.87}

\definecolor{cmarkgreen}{rgb}{0.216,0.545,0.18}
\definecolor{xmarkred}{rgb}{0.655,0.22,0.239}
\newcommand{\cmark}{{\color{cmarkgreen}\ding{51}}}%
\newcommand{\xmark}{{\color{xmarkred}\ding{55}}}%

\newcommand{\sparagraph}[1]{\noindent \textbf{#1}\ \ {}}

\newcommand{\nb}[1]{}

\title{The \bmname{} Benchmark for Robust Imitation}

\author{%
  Sam Toyer \qquad Rohin Shah \qquad Andrew Critch \qquad Stuart Russell\\
  Department of Electrical Engineering and Computer Sciences\\
  University of California, Berkeley\\
  \{\texttt{sdt,rohinmshah,critch,russell}\}\texttt{@berkeley.edu} \\
}

\begin{document}

\maketitle

\begin{abstract}
Imitation Learning (IL) algorithms are typically evaluated in the same environment that was used to create demonstrations.
This rewards precise reproduction of demonstrations in one particular environment, but provides little information about how robustly an algorithm can generalise the demonstrator's intent to substantially different deployment settings.
This paper presents the \bmname{} benchmark suite, which permits \textit{systematic} evaluation of generalisation by quantifying robustness to different kinds of distribution shift that an IL algorithm is likely to encounter in practice.
Using the \bmname{} suite, we confirm that existing IL algorithms overfit significantly to the context in which demonstrations are provided.
We also show that standard methods for reducing overfitting are effective at creating narrow perceptual invariances, but are not sufficient to enable transfer to contexts that require substantially different behaviour, which suggests that new approaches will be needed in order to robustly generalise demonstrator intent.
Code and data for the MAGICAL suite is available at \url{https://github.com/qxcv/magical/}.
\end{abstract}

\section{Introduction}\label{sec:intro}

Imitation Learning (IL) is a practical and accessible way of programming robots to perform useful tasks~\cite{billard2008survey}.
For instance, the owner of a new domestic robot might spend a few hours using tele-operation to complete various tasks around the home: doing laundry, watering the garden, feeding their pet salamander, and so on.
The robot could learn from these demonstrations to complete the tasks autonomously.
For IL algorithms to be useful, however, they must be able to learn how to perform tasks from few demonstrations.
A domestic robot wouldn't be very helpful if it required thirty demonstrations before it figured out that you are deliberately washing your purple cravat separately from your white breeches, or that it's important to drop bloodworms \textit{inside} the salamander tank rather than next to it.
Existing IL algorithms assume that the environment observed at test time will be identical to the environment observed at training time, and so they cannot generalise to this degree.
Instead, we would like algorithms that solve the task of \textit{robust IL}: given a small number of demonstrations in one training environment, the algorithm should be able to generalise the intent behind those demonstrations to (potentially very different) deployment environments.

One barrier to improved algorithms for robust IL is a lack of appropriate benchmarks.
IL algorithms are commonly tested on Reinforcement Learning (RL) benchmark tasks, such as those from OpenAI Gym~\cite{torabi2018behavioral,ho2016generative,kostrikov2018discriminator,brockman2016openai}.
However, the demonstrator intent in these benchmarks is often trivial (e.g.\ the goal for most of Gym's MuJoCo tasks is simply to run forward), and limited variation in the initial state distribution means that algorithms are effectively being evaluated in the same setting that was used to provide demonstrations.
Recent papers on Inverse Reinforcement Learning (IRL)---which is a form of IL that infers a reward under which the given demonstrations are near-optimal---have instead used ``testing'' variants of standard Gym tasks which differ from the original demonstration environment~\cite{fu2017learning,yu2019meta,peng2018variational,qureshi2018adversarial}.
For instance, \citet{fu2017learning} trained an algorithm on demonstrations from the standard ``Ant'' task from Gym, then tested on a variant of the task where two of the creature's four legs were disabled.
Splitting the environment into such ``training'' and ``test'' variants makes it possible to measure the degree to which an algorithm overfits to task-irrelevant features of the supplied demonstrations.
However, there is so far no standard benchmark for robust IL, and researchers must instead use ad-hoc adaptations of RL benchmarks---such as the modified Ant benchmark and similar alternatives discussed in \cref{sec:related}---to evaluate intent generalisation.

\begin{figure}[t]
    \centering
    \includegraphics{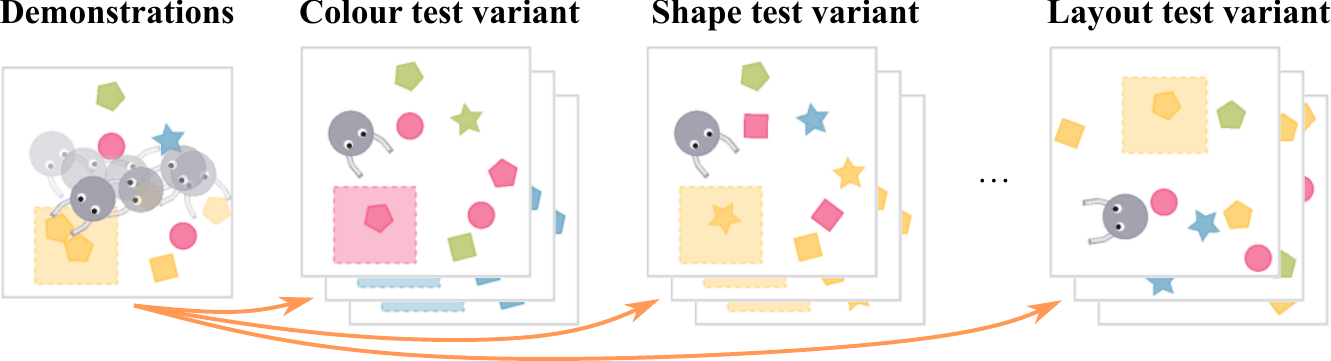}
    \caption{
        Unlike existing IL benchmarks, \bmname{} makes a distinction between \textit{demonstration} and \textit{test} variants of a task.
        Demonstrations are all provided in one particular configuration of the world (the ``demonstration variant'').
        The learnt policy (or reward function) is then evaluated across a set of \textit{test variants}, each of which randomise one aspect of the environment, such as block colour or shape, environment layout, dynamics, etc.
        This makes it possible to understand precisely which aspects of the underlying task the algorithm has been able to infer from demonstrations.
    }
    \label{fig:lead}
\end{figure}

To address the above issues, we introduce the Multitask Assessment of Generalisation in Imitative Control ALgorithms (\bmname{}).
Each \bmname{} task occurs in the same 2D ``\bmname{} universe'', where environments consist of a robot with a gripper surrounded by a variable number of objects in a fixed-size workspace.
Each task is associated with a \textit{demonstration variant}, which is a fixed initial state from which all human demonstrations are provided.
A task is also associated with a set of \textit{test variants} for which no demonstrations are provided.
As illustrated in \cref{fig:lead}, the test variants each randomise a different aspect of the world, such as object colour, transition dynamics, or object count.
Randomising attributes of objects and the physics of the world lets us evaluate the ability of a robust IL algorithm to perform \textit{combinatorial generalisation}~\cite{battaglia2018relational}.
For instance, given a demonstration of the robot pushing a red square across the workspace, an algorithm should be able to push a yellow circle across the workspace; given a demonstration of three green and yellow blocks being placed in a line, an algorithm should also be able to place four red and blue blocks in a line; and so on.
%

\bmname{} has several advantages over evaluation methods for standard (non-robust) IL:
\begin{itemize}
    \item \textbf{No ``training on the test set''.}
    Evaluating in the same setting that was used to give demonstrations allows algorithms to exploit features that might not be present during deployment.
    Having separate test variants for a task allows us to identify this kind of overfitting.

    \item \textbf{Distinguishes between different types of transfer.}
    Each test variant evaluates robustness to a distinct, semantically meaningful axis of variation.
    This makes it possible to characterise precisely which aspects of the provided demonstrations a given algorithm is relying on, and to diagnose the causes of over- or under-fitting.

    \item \textbf{Enables knowledge reuse between tasks.}
    Each \bmname{} task requires similar concepts and low-level skills to solve.
    Different tasks can therefore provide ``background knowledge'' for multi-task and meta-IL algorithms, such as knowledge that objects can have different colours, or that objects with different shapes respond in a particular way when grasped.
\end{itemize}
Our experiments in \cref{sec:expts} demonstrate the brittleness of standard IL algorithms, particularly under large shifts in object position or colour.
We also show that common methods for improving generalisation---such as multitask training, data augmentation, and alternative camera views---sometimes improve robustness to small changes, but still fail to generalise to more extreme ones.

\section{\bmname{}: Systematically evaluating robust IL}

We will now introduce the main elements of the \bmname{} benchmark.
We first describe the abstract setup of our benchmark, then detail the specific tasks and variants available in the benchmark.

\subsection{Benchmark setup}

The \bmname{} benchmark consists of a set of \textit{tasks} $\mathcal T_1, \mathcal T_2, \ldots, \mathcal T_m$.
Each task can in turn be broken down into \textit{variants} of a single base Markov Decision Process (MDP) that provide different state distributions and ``physics'' for an agent.
Formally, each task $\mathcal T = (S, v^D, \mathcal V)$ consists of a \textit{scoring function} $S(\tau)$, a \textit{demonstration variant} $v^D$, and a set of additional \textit{test variants} $\mathcal V = \{v_1, v_2, \ldots, v_n\}$.
The scoring function $S(\tau)$ takes a trajectory $\tau=(s_0, a_0, s_1, a_1, \ldots, s_T, a_T)$ and assigns it a score $S(\tau)\in [0,1]$, where 0 is the score of a no-op policy, and 1 is the score of a perfect demonstrator.
Unlike a reward function, $S(\tau)$ need not be Markovian.
%
%
In order to evaluate generalisation, the variants are split into a single demonstration variant $v^D$ and a set of test variants $\mathcal V$.

In our domestic robotics analogy, $v^D$ might represent a single room and time-of-day in which demonstrations for some domestic task collected, while each test variant $v \in \mathcal V$ could represent a different room, different time-of-day, and so on.
Algorithms are expected to be able to take demonstrations given only in demonstration variant $v^D$, then generalise the intent behind those demonstrations in order to achieve the same goal in each test variant $v \in \mathcal V$.
This can be viewed either as a form of domain transfer, or as ordinary generalisation using only a single sample from a hypothetical distribution over all possible variants of each task.

Formally, each variant associated with a task $\mathcal T$ defines a distribution over reward-free MDPs.
%
Specifically, a variant $v = (p_0, p_\rho, H)$ consists of an \textit{initial state distribution} $p_0(s_0)$, a \textit{dynamics distribution} $p_\rho(\rho)$, and a \textit{horizon} $H$.
States are fully observable via an image-based observation space.
Further, all variants have the same state space, the same observation space, and the same action space, which we discuss below.
In addition to sampling an initial state $s_0 \sim p_0(s_0)$, at the start of each trajectory, a \textit{dynamics vector} $\rho \in \mathbb R^d$ is also sampled from the dynamics distribution $p_\rho(\rho)$.
Unlike the state, \textit{$\rho$} is not observable to the agent; this vector controls aspects of the dynamics such as friction and motor strength.
Finally, the horizon $H$ defines a fixed length for all trajectories sampled from the MDP associated with the variant $v$.
Each variant associated with a given task has the same fixed horizon $H$ to avoid ``leaking'' information about the goal through early termination~\cite{kostrikov2018discriminator}.

All tasks and variants in the \bmname{} benchmark share a common continuous state space $\mathcal S$.
A state $s\in \mathcal S$ consists of a configuration (pose, velocity, and gripper state) $q_R$ for the robot, along with object configurations $\mathcal O = \{o_1, \ldots, o_E\}$ (where the number of objects in $s_0$ may be random).
In addition to pose, each object configuration $o_i$ includes an object \textit{type} and a number of fixed attributes.
Objects can be of two types: \textit{blocks} (small shapes that can be pushed around by the agent) and \textit{goal regions} (coloured rectangles that the agent can drive over, but not push around).
Each block has a fixed shape (square, pentagon, star, or circle) and colour (red, green, blue, or yellow).
Each goal region has a fixed colour, width, and height.
In order to facilitate generalisation across tasks with a different number of objects, we use a common image-based observation space and discrete, low-level action space for all tasks, which we describe in detail in \cref{app:benchmark-obs-act}.
At an implementation level, we expose each variant of each task as a distinct Gym environment~\cite{brockman2016openai}, which makes it straightforward to incorporate MAGICAL into existing IL and RL codebases.

\subsection{Tasks and variants}\label{ssec:tasks-variants}

With the handful of building blocks listed in the previous section,
we can create a wide variety of tasks, which we describe in \cref{ssec:tasks}.
The object-based structure of the environment also makes it easy to evaluate combinatorial generalisation by randomising one or more attributes of each object while keeping the others fixed, as described in \cref{ssec:variants}.
%
%

\subsubsection{Tasks}\label{ssec:tasks}

\begin{figure}
    \newcommand{\dpicw}{0.22\textwidth} 
    \centering
    \begin{subfigure}{\dpicw}
        \centering
        \includegraphics[width=\textwidth]{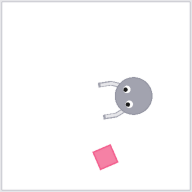}
        \caption{MoveToCorner}
    \end{subfigure}
    \hfill
    \begin{subfigure}{\dpicw}
        \centering
        \includegraphics[width=\textwidth]{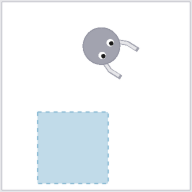}
        \caption{MoveToRegion}
    \end{subfigure}
    \hfill
    \begin{subfigure}{\dpicw}
        \centering
        \includegraphics[width=\textwidth]{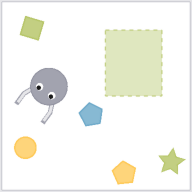}
        \caption{MatchRegions}
    \end{subfigure}
    \hfill
    \begin{subfigure}{\dpicw}
        \centering
        \includegraphics[width=\textwidth]{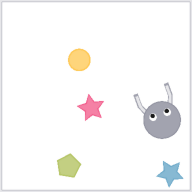}
        \caption{MakeLine}
    \end{subfigure}
    \\[2mm]
    
    \begin{subfigure}{\dpicw}
        \centering
        \includegraphics[width=\textwidth]{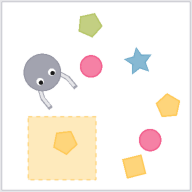}
        \caption{FindDupe}
    \end{subfigure}
    \hfill
    \begin{subfigure}{\dpicw}
        \centering
        \includegraphics[width=\textwidth]{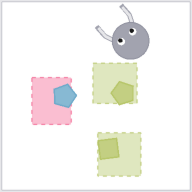}
        \caption{FixColour}
    \end{subfigure}
    \hfill
    \begin{subfigure}{\dpicw}
        \centering
        \includegraphics[width=\textwidth]{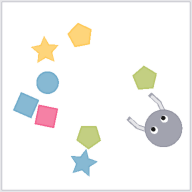}
        \caption{ClusterColour}
    \end{subfigure}
    \hfill
    \begin{subfigure}{\dpicw}
        \centering
        \includegraphics[width=\textwidth]{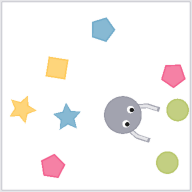}
        \caption{ClusterShape}
    \end{subfigure}
    \caption{Demonstration variants for \bmname{} tasks.
    \cref{app:benchmark} shows an example demonstration for each task.}
    \label{fig:demo-variants}
\end{figure}

Tasks in the \bmname{} suite were chosen to balance three desiderata.
First, given a handful of trajectories from the demonstration variant of a task, it should be possible for a human observer to infer the goal with sufficient accuracy to solve the test variants.
We have chosen demonstration variants (illustrated in \cref{fig:demo-variants}) that rule out obvious misinterpretations, like mistakenly identifying colour as being task-relevant when it is not.
Second, the tasks should be constructed so that they involve complementary skills that meta- and multi-task learning algorithms can take advantage of.
In our tasks, these ``shared skills'' include block manipulation; identification of colour or shape; and relational reasoning.
Third, the demonstration variant of each task must be solvable by existing (non-robust) IL algorithms.
This ensures that the main challenge of the MAGICAL suite lies in \textit{generalising} to the test variants (robust IL), as opposed to reproducing the demonstrator's behaviour in the demonstration variant (standard IL).
This section briefly describes the resulting tasks; detailed discussion of horizons, score functions, etc.\ is deferred to \cref{app:benchmark}.

\newcommand{\tpicw}{0.18\textwidth} 
\newcommand{\taskpic}[1]{
}


\begingroup

\taskpic{figures/tasks/static-movetocorner-demo-v0.png}
\sparagraph{Move to Corner (MTC)} The robot must push a single block from one corner of the workspace to the diagonally opposite corner.
%
%
Test variants are constrained so that the robot and block start near the lower right corner.
The score is $S(\tau) = 1$ if the block finishes the trajectory in the top left eighth of the workspace, and decreases to zero as the block gets further from the top left corner.
\nb{\ST{Space permitting, try to explain what each of these tasks contributes to the whole set. What do they evaluate that other tasks do not evaluate? Are they meant to be easy or hard? etc.}}

\endgroup


\begingroup

\taskpic{figures/tasks/static-movetoregion-demo-v0.png}
\sparagraph{MoveToRegion (MTR)} The robot must drive inside a goal region and stay there.
There are no blocks in the demonstration or test variants.
Further, variants only have one goal region to ensure that the objective is unambiguous.
The agent's score is $S(\tau)=1$ if the robot's body is inside the goal region at the end of the trajectory, and $S(\tau) = 0$ otherwise.

\endgroup


\begingroup

\taskpic{figures/tasks/static-matchregions-demo-v0.png}
\sparagraph{MatchRegions (MR)} There is a set of coloured blocks and a goal region visible to the robot, and the robot must push all blocks of the same colour as the goal region into the goal region.
%
%
Test variants are constrained to have one goal region and at least one block of the same colour as that goal region.
A perfect score is given upon termination if the goal regions contains all and only blocks of the goal region's colour, with penalties for excluding any blocks of the goal colour, or including other blocks.

\endgroup


\begingroup

\taskpic{figures/tasks/static-makeline-demo-v0.png}
\sparagraph{MakeLine (ML)} Here the objective is for the robot to arrange all the blocks in the workspace into a single line.
A perfect score is given if all blocks are approximately colinear and close together; a penalty is given for each block that does not form part of the longest identifiable line.
Refer to \cref{app:benchmark} for details on how a ``line'' is defined.

\endgroup


\begingroup

\taskpic{figures/tasks/static-finddupe-demo-v0.png}
\sparagraph{FindDupe (FD)} Similar to MatchRegions, except the goal region initially contains a ``query'' block which has the same shape and colour as at least one other block outside the goal region.
%
The objective is to push at least one of those duplicate blocks into the goal region, which yields a perfect score.
Penalties are given for knocking the query block out of the goal region, failing to find a duplicate, or pushing non-duplicate blocks into the goal region.

\endgroup


\begingroup

\taskpic{figures/tasks/static-fixcolour-demo-v0.png}
\sparagraph{FixColour (FC)} In each variant of this task, the workspace contains a set of non-overlapping goal regions.
Each goal region contains a single block, and exactly one block in the workspace will have a different colour to its enclosing goal region.
A perfect score is given for pushing that block out of its enclosing goal region and into an unoccupied part of the workspace, without disturbing other blocks.
%

\endgroup


\begingroup

\taskpic{figures/tasks/static-clustercolour-demo-v0.png}
\sparagraph{ClusterColour (CC) and ClusterShape (CS)} The robot is confronted with a jumble of blocks of different colours and shapes.
It must push the blocks into clusters of either uniform colour (in the CC task), or uniform shape (in the CS task).
Test variants are constrained to include at least one block of each colour and each shape.
A perfect score is given for creating four spatially distinct clusters corresponding to each of the four colours (CC) or shapes (CS), with a penalty proportional to the number of blocks that do not belong to an identifiable cluster.

\endgroup


\subsubsection{Test variants}\label{ssec:variants}

In addition to its demonstration variant, each of the tasks above has a set of associated test variants.
%
%
Some variants are not supported for tasks that do not have any blocks, or where the initial state is otherwise restricted, as documented in \cref{tab:variants} of \cref{app:benchmark}.
\begin{description}
\item[Jitter] Takes demo variant and randomly perturbs the poses of the robot and all objects by up to 5\% of the maximum possible range.
Failure on this variant indicates severe overfitting to the demonstration variant (e.g.\ by memorising action sequences).

\item[Layout] Completely randomises the position and orientation of the robot and all blocks, plus position and dimensions of goal regions; a more challenging version of Jitter.

\item[Colour] Block colours are randomly reassigned as appropriate for the task.
This tests whether the agent is responsive to block colour (when it is task-relevant, like in CC and MR), or is correctly ignorant of colour (when it is irrelevant, like in MTC and CS).

\item[Shape] Similar to Colour, except the shapes of blocks are randomised rather than the colours.
This variant either tests for appropriate responsiveness or invariance to shape, depending on whether shape is task-relevant.

\item[CountPlus] The number of blocks is randomised (along with shape, colour, and position) to test whether the agent can handle ``larger'' or ``smaller'' problems (i.e.\ ``generalisation to $n$''~\cite{shavlik1990acquiring}).

\item[Dynamics] Subtly randomises friction of objects and the robot against the workspace, as well as force of robot motors (for rotation, forward/backward motion, and the gripper).

\item[All] Combines all applicable variants for a task (e.g.\ Layout, Colour, Shape, CountPlus, Dynamics).
\end{description}

\section{Data-efficient intent disambiguation}\label{sec:algos}

Succeeding at the \bmname{} benchmark requires agents to generalise the intent behind a set of demonstrations to substantially different test variants.
%
%
%
%
%
%
We anticipate that resolving the ambiguity inherent in this task will require additional sources of information about the demonstrator's goal beyond just single-task demonstrations.
In this section, we review two popular non-robust IL algorithms, as well as some common ways in which alternative sources of goal information are incorporated into these algorithms to improve generalisation.
%

\subsection{Baseline methods}\label{ssec:algos-base-methods}

Our first baseline method is Behavioural Cloning (BC).
BC treats a demonstration dataset $\mathcal D$ as an undistinguished collection of state-action pairs $\{(s_1, a_1), \ldots, (s_M, a_M)\}$.
It then optimises the parameters $\theta$ of the policy $\pi_\theta(a \mid s)$ via gradient descent on the log loss
\begin{equation*}
    \mathcal L_\bc(\theta; \mathcal D) = -\expect_{\mathcal D} \log \pi_\theta(a \mid s)~.
\end{equation*}

Our second baseline method is Generative Adversarial IL (GAIL)~\cite{ho2016generative}.
GAIL casts IL as a GAN problem~\cite{goodfellow2014generative}, where the generator $\pi_\theta(a \mid s)$ is an imitation policy, and the discriminator $D_\psi : \mathcal S \times \mathcal A \to [0,1]$ is tasked with distinguishing imitation behaviour from expert behaviour.
Specifically, GAIL uses alternating gradient descent to approximate a saddle point of
\begin{align*}
    \max_\theta \min_\psi \left\{\mathcal L_\adv(\theta, \psi; \mathcal D) = -\expect_{\pi_\theta} \log D_\psi(s,a) - \expect_{\mathcal D} \log(1 - D_\psi(s,a)) + \lambda H(\pi_\theta)\right\}~,
\end{align*}
where $H$ denotes entropy and $\lambda \geq 0$ is a policy regularisation parameter.
%

We also included a slight variation on GAIL which (approximately) minimises Wasserstein divergence between occupancy measures, rather than Jensen-Shannon divergence.
We refer to this baseline as WGAIL-GP.
In analogy with WGAN-GP~\cite{gulrajani2017improved}, WGAIL-GP optimises the cost
\begin{align*}
    \max_\theta \min_\psi \left\{\mathcal L_{\text{w-gp}}(\theta, \psi; \mathcal D) = \expect_{\mathcal D} D_\psi(s,a) - \expect_{\pi_\theta} D_\psi(s,a) + \lambda_{\text{w-gp}} \expect_{\frac{1}{2} \pi_\theta + \frac{1}{2} \mathcal D} (\|\nabla_s D(s,a)\|_2 - 1)^2 \right\}~,
\end{align*}
The gradient penalty approximately enforces 1-Lipschitzness of the discriminator by encouraging the norm of the gradient to be 1 at points between the support of $\pi_\theta$ and $\mathcal D$.
Since actions were discrete, we did not enforce 1-Lipschitzness with respect to the action input.
We also did not backpropagate gradients with respect to the gradient penalty back into the policy parameters $\theta$, since the gradient penalty is only intended as a soft constraint on $D$.

In addition to these baselines, we also experimented with Apprenticeship Learning (AL).
Unfortunately we could not get AL to perform well on most of our tasks, so we defer further discussion of AL to \cref{app:expt-details}.

\subsection{Using multi-task data}\label{ssec:algos-multi-task}

As noted earlier, the \bmname{} benchmark tasks have similar structure, and should in principle benefit from multi-task learning.
Specifically, say we are given a multi-task dataset $\mathcal D_{\rm mt} = \{\mathcal D(\mathcal T_i, v_i^D, n_i)\}_{i=1}^M$, where $\mathcal D(\mathcal T_i, v, n)$ denotes a dataset of $n$ trajectories for variant $v$ of task $\mathcal T_i$.
For BC and GAIL, we can decompose the policy for task $\mathcal T_i$ as $\pi_{\theta}^i = g^i_\theta \circ f_\theta$, where $f_\theta : \mathcal S \to \mathbb R^d$ is a multi-task state encoder, while $g^i_\theta : \mathbb R^d \to \Delta(\mathcal A)$ is a task-specific policy decoder.
We can also decompose the GAIL discriminator as $D^i_\psi = s^i_\psi \circ r_\psi$, where $r_\psi : \mathbb S \times \mathbb A \to \mathbb R^d$ is shared and $s^i_\psi: \mathbb R^d \to [0,1]$ is task-specific.
We then modify the BC and GAIL objectives to
\begin{align*}
    \mathcal L_\bc(\theta; \mathcal D_{\mt})
    &=
    \sum_{i=1}^M \mathcal L_\bc(\theta; \mathcal D(\mathcal T_i, v_i^D, n_i))
    \ \ \text{and}\ \ 
    \mathcal L_\adv(\theta, \psi; \mathcal D_{\mt})
   =
    \sum_{i=1}^M \mathcal L_\adv(\theta, \psi; \mathcal D(\mathcal T_i, v_i^D, n_i))~.
\end{align*}

\subsection{Domain-specific priors and biases}\label{ssec:algos-biases}

Often the most straightforward way to improve the robustness of an IL algorithm is to constrain the solution space to exclude common failure modes.
For instance, one could use a featurisation that only captures task-relevant aspects of the state.
Such priors and biases are generally domain-specific; for the image-based \bmname{} suite, we investigated two such biases:
\begin{itemize}
    \item \textbf{Data augmentation:}
    In \bmname{}, our score functions are invariant to whether objects are repositioned or rotated slightly; further, human observers are typically invariant to small changes in colour or local image detail.
    As such, we used random rotation and translation, Gaussian noise, and colour jitter to augment training data for the BC policy and GAIL discriminator.
    This can be viewed as a post-hoc form of domain randomisation, which has previously yielded impressive results in robotics and RL~\cite{akkaya2019solving}.
    We found that GAIL discriminator augmentations were necessary for the algorithm to solve more-challenging tasks, as previously observed by \citet{zolna2019task}.
    In BC, we found that policy augmentations improved performance on both demonstration and test variants.

    \item \textbf{Ego- and allocentric views:}
    Except where indicated otherwise, all of the experiments in \cref{sec:expts} use an \textit{egocentric} perspective, which always places the agent at the same position (and in the same orientation) within the agent's field of view.
    This contrasts with an \textit{allocentric} perspective, where observations are focused on a fixed region of the environment (in our case, the extent of the workspace), rather than following the agent's position.
    %
    %
    In the context of language-guided visual navigation, \citet{hill2020environmental} previously found that an egocentric view improved generalisation to unseen instructions or unseen visual objects, despite the fact that it introduces a degree of partial observability to the environment.
    %
    %
\end{itemize}

\section{Experiments}\label{sec:expts}

Our empirical evaluation has two aims.
First, to confirm that single-task IL methods fail to generalise beyond the demonstration variant in the \bmname{} suite.
Second, to analyse the ways in which the common modifications discussed in \cref{sec:algos} affect generalisation.

\subsection{Experiment details}

We evaluated all the single- and multi-task algorithms in \cref{sec:algos}, plus augmentation and perspective ablations, on all tasks and variants.
Each algorithm was trained five times on each task with different random seeds.
In each run, the training dataset for each task consisted of 10 trajectories from the demo variant.
%
%
All policies, value functions, and discriminators were represented by Convolutional Neural Networks (CNNs).
Observations were preprocessed by stacking four temporally adjacent RGB frames and resizing them to 96$\times$96 pixels.
%
%
For multi-task experiments, task-specific weights were used for the final fully-connected layer of each policy/value/discriminator network, but weights of all preceding layers were shared.
%
%
%
The BC policy and GAIL discriminator both used translation, rotation, colour jitter, and Gaussian noise augmentations by default.
The GAIL policy and value function did not use augmented data, which we found made training unstable.
Complete hyperparameters and data collection details are listed in \cref{app:expt-details}.
The IL algorithm implementations that we used to generate these results are available on GitHub,\footnote{Multi-task imitation learning algorithms: \url{https://github.com/qxcv/mtil/}} as is the MAGICAL benchmark suite and all demonstration data.\footnote{Benchmark suite and links to data: \url{https://github.com/qxcv/magical/}}

\subsection{Discussion}

Due to space limitations, this section addresses only a selection of salient patterns in the results.
\cref{tab:short-results} provides score statistics for a subset of algorithms and variants, averaged across \textit{all} tasks.
See \cref{ssec:tasks} for task name abbreviations (MTR, FC, etc.).
Because the tasks vary in difficulty, pooling across all tasks yields high score variance in \cref{tab:short-results}.
Actual score variance for each method is much lower when results are constrained to just one task; refer to \cref{app:results} for complete results.

\sparagraph{Overfitting to position}
All algorithms exhibited severe overfitting to the \textit{position} of objects.
The Layout, CountPlus, and All variants yielded near-zero scores in all tasks except MTC and MTR, and on many tasks there was also poor transfer to the Jitter variant.
\nb{\ST{Could include a picture showing the agent making the exact same motion even after an object is moved (e.g. in a Jitter variant).}}
For some tasks, we found that the agent would simply execute the same motion regardless of its initial location or the positions of task-relevant objects.
This was true on the FC task, where the agent would always execute a similar forward arc regardless of its initial position, and also noticeable on MTC and FD, where the agent would sometimes move to the side of a desired block when it was shifted slightly.
For BC, this issue was ameliorated by the use of translation and rotation augmentations, presumably because the policy could better handle small deviations from the motions seen at training time.

\begin{table}[t]
    \centering
        \begin{tabular}{lccccc}
\toprule
{Method} &                                           Demo &                                         Jitter &                                         Layout &                                         Colour &                                          Shape \\
\midrule
{BC (single-task)                     } &  \cellcolor[rgb]{0.85,0.91,0.95} 0.64$\pm$0.29 &  \cellcolor[rgb]{0.87,0.93,0.96} 0.56$\pm$0.27 &  \cellcolor[rgb]{0.97,0.98,0.99} 0.14$\pm$0.16 &  \cellcolor[rgb]{0.92,0.95,0.98} 0.39$\pm$0.30 &  \cellcolor[rgb]{0.88,0.93,0.97} 0.52$\pm$0.33 \\
{\quad{}Allocentric             } &  \cellcolor[rgb]{0.86,0.92,0.96} 0.58$\pm$0.33 &  \cellcolor[rgb]{0.89,0.94,0.97} 0.48$\pm$0.29 &  \cellcolor[rgb]{0.99,0.99,1.00} 0.04$\pm$0.04 &  \cellcolor[rgb]{0.91,0.95,0.97} 0.42$\pm$0.32 &  \cellcolor[rgb]{0.88,0.94,0.97} 0.50$\pm$0.37 \\
{\quad{}No augmentations            } &  \cellcolor[rgb]{0.87,0.93,0.96} 0.55$\pm$0.37 &  \cellcolor[rgb]{0.92,0.96,0.98} 0.37$\pm$0.30 &  \cellcolor[rgb]{0.97,0.98,0.99} 0.12$\pm$0.15 &  \cellcolor[rgb]{0.94,0.96,0.98} 0.33$\pm$0.30 &  \cellcolor[rgb]{0.91,0.95,0.97} 0.41$\pm$0.33 \\
{\quad{}No trans./rot. aug.} &  \cellcolor[rgb]{0.87,0.93,0.96} 0.55$\pm$0.37 &  \cellcolor[rgb]{0.91,0.95,0.97} 0.41$\pm$0.31 &  \cellcolor[rgb]{0.97,0.98,0.99} 0.13$\pm$0.15 &  \cellcolor[rgb]{0.93,0.96,0.98} 0.33$\pm$0.30 &  \cellcolor[rgb]{0.91,0.95,0.97} 0.43$\pm$0.35 \\
{\quad{}Multi-task                     } &  \cellcolor[rgb]{0.86,0.92,0.96} 0.59$\pm$0.33 &  \cellcolor[rgb]{0.88,0.93,0.96} 0.53$\pm$0.31 &  \cellcolor[rgb]{0.97,0.98,0.99} 0.14$\pm$0.18 &  \cellcolor[rgb]{0.94,0.97,0.98} 0.30$\pm$0.25 &  \cellcolor[rgb]{0.88,0.93,0.97} 0.51$\pm$0.36 \\
{GAIL (single-task)                   } &  \cellcolor[rgb]{0.82,0.90,0.96} 0.72$\pm$0.35 &  \cellcolor[rgb]{0.83,0.90,0.96} 0.69$\pm$0.33 &  \cellcolor[rgb]{0.96,0.97,0.99} 0.22$\pm$0.23 &  \cellcolor[rgb]{0.95,0.97,0.99} 0.27$\pm$0.24 &  \cellcolor[rgb]{0.86,0.92,0.96} 0.60$\pm$0.42 \\
{\quad{}Allocentric            } &  \cellcolor[rgb]{0.87,0.92,0.96} 0.57$\pm$0.46 &  \cellcolor[rgb]{0.89,0.94,0.97} 0.49$\pm$0.40 &  \cellcolor[rgb]{0.99,0.99,1.00} 0.03$\pm$0.03 &  \cellcolor[rgb]{0.92,0.96,0.98} 0.39$\pm$0.36 &  \cellcolor[rgb]{0.88,0.94,0.97} 0.50$\pm$0.45 \\
{\quad{}No augmentations          } &  \cellcolor[rgb]{0.90,0.95,0.97} 0.44$\pm$0.42 &  \cellcolor[rgb]{0.94,0.96,0.98} 0.32$\pm$0.31 &  \cellcolor[rgb]{0.98,0.99,1.00} 0.09$\pm$0.12 &  \cellcolor[rgb]{0.96,0.98,0.99} 0.19$\pm$0.23 &  \cellcolor[rgb]{0.95,0.97,0.98} 0.28$\pm$0.33 \\
{\quad{}WGAIL-GP               } &  \cellcolor[rgb]{0.91,0.95,0.97} 0.42$\pm$0.38 &  \cellcolor[rgb]{0.93,0.96,0.98} 0.33$\pm$0.32 &  \cellcolor[rgb]{0.97,0.98,0.99} 0.14$\pm$0.20 &  \cellcolor[rgb]{0.98,0.99,1.00} 0.10$\pm$0.11 &  \cellcolor[rgb]{0.94,0.96,0.98} 0.33$\pm$0.33 \\
{\quad{}Multi-task                   } &  \cellcolor[rgb]{0.93,0.96,0.98} 0.37$\pm$0.41 &  \cellcolor[rgb]{0.94,0.96,0.98} 0.33$\pm$0.36 &  \cellcolor[rgb]{0.97,0.98,0.99} 0.16$\pm$0.25 &  \cellcolor[rgb]{0.98,0.99,0.99} 0.11$\pm$0.12 &  \cellcolor[rgb]{0.95,0.97,0.99} 0.28$\pm$0.36 \\
\bottomrule
\end{tabular}
    \vspace{2mm}
    \caption{
    Score statistics for a subset of variants and compared algorithms.
    We report the mean and standard deviation of test scores aggregated across \textit{all} tasks, with five seeds per algorithm and task.
    Darker colours indicate higher scores.}
    \label{tab:short-results}
\end{table}

\sparagraph{Colour and shape transfer}
Surprisingly, BC and GAIL both struggled with colour transfer to a greater degree than shape transfer on several tasks, as evidenced by the aggregated statistics for Colour and Shape variants in \cref{tab:short-results}.
%
%
Common failure modes included freezing in place or moving in the wrong direction when confronted with an object of a different colour to that seen at training time.
In contrast, in most tasks where shape invariance was desirable (including MTC, MR, ML, and FC), the agent had no trouble reaching and manipulating blocks of different shapes.
Although colour jitter was one of the default augmentations, the BC ablations in \cref{tab:short-results} suggest that almost all of the advantage of augmentations comes from the use of translation/rotation augmentations.
In particular, we did not find that colour jitter greatly improved performance on tasks where the optimal policy was colour-invariant.
In spite of exposing the networks to a greater range of colours at train time, multitask training also failed to improve colour transfer, as we discuss below.
Although translation and rotation sometimes improved colour transfer (e.g. for BC on FindDupe in \cref{tab:results-tab3}), it is not clear why this was the case.
We speculate that these augmentations could have encouraged the policy to acquire more robust early-layer features for edge and corner detection that did not rely on just one colour channel.

\sparagraph{Multi-task transfer}
Plain multi-task learning had mixed effects on generalisation.
In some cases it improved generalisation (e.g.\ for BC on FC), but in most cases it led to unchanged or \textit{negative} transfer, as in the Colour test variants for MTC, MR, and FD.
%
%
%
This could have been because the policy was using colour to distinguish between tasks.
More speculatively, it may be that a multi-task BC or GAIL loss is not the best way to incorporate off-task data, and that different kinds of multi-task pretraining are necessary (e.g. learning forward or inverse dynamics~\cite{brown2020safe}).
%
%
%

\sparagraph{Egocentric view and generalisation}
\nb{\ST{A picture illustrating ego generalisation MTR might be useful as a prototypical example.}}
The use of an allocentric (rather than egocentric) view did not improve generalisation or demo variant performance for most tasks, and sometimes decreased it.
\cref{tab:short-results} shows the greatest performance drop on variants that change object position, such as Layout and Jitter.
For example, in MTR we found that egocentric policies tended to rotate in one direction until the goal region was in the centre of the agent's field of view, then moved forward to reach the region, which generalises well to different goal region positions.
In contrast, the allocentric policy would often spin in place or get stuck in a corner when confronted with a goal region in a different position.
This supports the hypothesis of \citet{hill2020environmental} that the egocentric view improves generalisation by creating positional invariances, and reinforces the value of being able to independently measure generalisation across distinct axes of variation (position, shape, colour, etc.).

\section{Related work}\label{sec:related}

There are few existing benchmarks that specifically examine robust IL.
The most similar benchmarks to \bmname{} have appeared alongside evaluations of IRL and meta-IL algorithms.
As noted in \cref{sec:intro}, several past papers employ ``test'' variants of standard Gym MuJoCo environments to evaluate IRL generalisation~\cite{fu2017learning,yu2019meta,peng2018variational,qureshi2018adversarial}, but these modified environments tend to have trivial reward functions (e.g.\ ``run forward'') and do not easily permit cross-environments transfer.
\citet{xu2018learning} and \citet{gleave2018multi} use gridworld benchmarks to evaluate meta- and multi-task IRL, and both benchmarks draw a distinction between demonstration and execution environments within a meta-testing task.
This distinction is similar in spirit to the demonstration/test variant split in \bmname{}, although \bmname{} differs in that it has more complex tasks and the ability to evaluate generalisation across different axes.
We note that there also exist dedicated IL benchmarks~\cite{memmesheimer2019simitate,james2019rlbench}, but they are aimed at solving challenging robotics tasks rather than evaluating generalisation directly.
%

There are many machine learning benchmarks that evaluate generalisation outside of IL.
For instance, there are several challenging benchmarks for generalisation~\cite{nichol2018gotta,cobbe2018quantifying,cobbe2019leveraging} and meta- or multi-task learning~\cite{yu2020meta} in RL.
Unlike \bmname{}, these RL benchmarks have no ambiguity about what the goal is in the training environment, since it is directly specified via a reward function.
Rather, the challenge is to ensure that the learnt policy (for model-free methods) can achieve that clearly-specified goal in different contexts (RL generalisation), or solve multiple tasks simultaneously (multi-task RL), or be adapted to new tasks with few rollouts (meta-RL).
There are also several instruction-following benchmarks for evaluating generalisation in natural language understanding~\cite{lake2018generalization,ruis2020benchmark}.
%
%
Although these are not IL benchmarks, they are similar to \bmname{} in that they include train/test splits that systematically evaluate different aspects of generalisation.
Finally, the Abstract Reasoning Corpus (ARC) is a benchmark that evaluates the ability of supervised learning algorithms to extrapolate geometric patterns in a human-like way~\cite{chollet2019measure}.
Although there is no sequential decision-making aspect to ARC, \citet{chollet2019measure} claims that solving the corpus may still require priors for ``objectness'', goal-directedness, and various geometric concepts, which means that methods suitable for solving \bmname{} may also be useful on ARC, and vice versa.

%
%
%

Although we covered some simple methods of improving IL robustness in \cref{sec:algos}, there also exist more sophisticated methods tailored to different IL settings.
Meta-IL~\cite{duan2017one,james2018task} and meta-IRL~\cite{xu2018learning,yu2019meta} algorithms assume that a large body of demonstrations is available for some set of ``train tasks'', but only a few demonstrations are available for ``test tasks'' that might be encountered in the future.
Each test task is assumed to have a distinct objective, but one that shares similarities with the train tasks, making it possible to transfer knowledge between the two.
These methods are likely useful for multi-task learning in the context of \bmname{}, too.
However, it's worth noting that past meta-IL work generally assumes that meta-train and meta-test settings are similar, whereas this work is concerned with how to generalise the intent behind a few demonstrations given in one setting (the demo variant) to other, potentially very different settings (the test variants).
Similar comments apply to existing work on multi-task IL and IRL~\cite{gleave2018multi,choi2012nonparametric,dimitrakakis2011bayesian,babes2011apprenticeship}.

%
%
%
%

\section{Conclusion}

\nb{\ST{Going to modify this conclusion to talk about how humans \textit{can} generalise in this way, and therefore we should expect IL algorithms to be able to do so as well. (need to skim some more papers on this first, though)}}
%
%
In this paper, we introduced the \bmname{} benchmark suite, which is the first imitation learning benchmark capable of evaluating generalisation across distinct, semantically-meaningful axes of variation in the environment.
Unsurprisingly, results for the \bmname{} suite confirm that single-task methods fail to transfer to changes in the colour, shape, position and number of objects.
However, we also showed that image augmentations and perspective shifts only slightly ameliorate this problem, and multi-task training can sometimes make it \textit{worse}.
This lack of generalisation stands in marked contrast to human imitation: even 14-month-old infants have been observed to generalise demonstrations of object manipulation tasks across changes in object colour and shape, or in the appearance of the surrounding room~\cite{barnat1996deferred}.
%
%
Closing the gap between current IL capabilities and human-like few-shot imitation could require significant innovations in multi-task learning, action and state representations, or models of human cognition.
The \bmname{} suite provides a way of evaluating such algorithms which not only tests whether they generalise well ``on average'', but also shines a light on the specific kinds of generalisation which they enable.

\section{Broader impact}

This paper presents a new benchmark for robust IL and argues for an increased focus on algorithms that can generalise demonstrator intent across different settings.
We foresee several possible follow-on effects from improved IL robustness:
\begin{description}
\item[Economic effects of automation] 
Better IL generalisation could allow for increased automation in some sectors of the economy.
This has the positive flow-on effect of increased economic productivity, but could lead to socially disruptive job loss.
Because our benchmark focuses on robust IL in robotics-like environments, it's likely that any effect on employment would be concentrated in sectors involving activities that are expensive to record.
This could include tasks like surgery (where few demonstrators are qualified to perform the task, and privacy considerations make it difficult to collect data) or packaging retail goods for postage (where few-shot learning might be important when there are many different types of goods to handle).

\item[Identity theft and model extraction]
More robust IL could enable better imitation of \textit{specific} people, and not just imitation of people in general.
This could lead to identify theft, for instance by mimicking somebody's speech or writing, or by fooling biometric systems.
Because this benchmark focuses on control and manipulation rather than media synthesis, it's unlikely that algorithms designed to solve our benchmark will be immediately useful for this purpose.
On the other hand, this concern is still relevant when applied to machine behaviour, rather than human behaviour.
In NLP, it's known that weights for ML models can be ``stolen'' by observing the model's outputs for certain carefully chosen inputs~\cite{krishna2020thieves}.
Similarly, more robust IL could make it possible to clone a robot's policy by observing its behaviour, which could make it harder to sell robot control algorithms as standalone products.

\item[Learnt objectives]
\citet{hadfield2016cooperative} argues that it is desirable for AI systems to infer their objectives from human behaviour, rather than taking them as fixed.
This can avoid problems that arise when an agent (human, robot, or organisation) doggedly pursues an easy-to-measure but incorrect objective, such as a corporate executive optimising for quarterly profit (which is easy to measure) over long-term profitability (which is actually desired by shareholders).
IL makes it possible to learn objectives from observed human behaviour, and more robust IL may therefore lead to AI systems that better serve their designers' goals.
However, it's worth noting that unlike, say, HAMDPs~\cite{fern2007decision} or CIRL games~\cite{hadfield2016cooperative}, IL cannot request clarification from a demonstrator if the supplied demonstrations are ambiguous, which limits its ability to learn the right objective in general.
Nevertheless, we hope that insights from improved IL algorithms will still be applicable to such interactive systems.
\end{description}

\begin{ack}
We would like to thank reviewers for helping to improve the presentation of the paper (in particular, clarifying the distinction between traditional IL and robust IL), and for suggesting additional related work and baselines.
This work was supported by a Berkeley Fellowship and a grant from the Open Philanthropy Project.
\end{ack}

\medskip
\bibliographystyle{plainnat}
\bibliography{citations}

\newpage
\appendix
\section{Additional benchmark details}\label{app:benchmark}

In this section we provide more details about our benchmark tasks, including horizons, scoring functions, and so on.
We also list the test variants available for each task in \cref{tab:variants}.

\begin{table}[H]
    \centering
    \begin{tabular}{@{}c@{\hskip 5mm}ccccccc@{}}
    \toprule
         \multirow{2}{*}{\textbf{Task}} & \multicolumn{7}{c}{\textbf{Test variant}}\\
         \cmidrule{2-8}
         & {Jitter} & {Layout} & {Colour} & {Shape} & {CountPlus} & {Dynamics} & {All}\\
         \midrule
         MoveToCorner & \cmark{} & \xmark{} & \cmark{} & \cmark{} & \xmark{} & \cmark{} & \cmark{}\\
         MoveToRegion & \cmark{} & \cmark{} & \cmark{} & \xmark{} & \xmark{} & \cmark{} & \cmark{}\\
         MatchRegions & \cmark{} & \cmark{} & \cmark{} & \cmark{} & \cmark{} & \cmark{} & \cmark{}\\
         MakeLine & \cmark{} & \cmark{} & \cmark{} & \cmark{} & \cmark{} & \cmark{} & \cmark{}\\
         FindDupe & \cmark{} & \cmark{} & \cmark{} & \cmark{} & \cmark{} & \cmark{} & \cmark{}\\
         FixColour & \cmark{} & \cmark{} & \cmark{} & \cmark{} & \cmark{} & \cmark{} & \cmark{}\\
         ClusterColour & \cmark{} & \cmark{} & \cmark{} & \cmark{} & \cmark{} & \cmark{} & \cmark{}\\
         ClusterType & \cmark{} & \cmark{} & \cmark{} & \cmark{} & \cmark{} & \cmark{} & \cmark{}\\
    \bottomrule
    \end{tabular}
    \vspace{2mm}
    \caption{Available variants for each task.
    Some variants are not defined for certain tasks because they may make task completion impossible, make task completion trivial (i.e.\ the null policy often completes the task), or do not provide a meaningful axis of variation (e.g.\ MoveToRegion does not feature any blocks, and so there are no shapes to randomise).}
    \label{tab:variants}
\end{table}

\subsection{Action and observation space}\label{app:benchmark-obs-act}

\begin{figure}[H]
    \centering
    \begin{subfigure}[b]{0.45\textwidth}
        \centering
        \includegraphics[width=\textwidth]{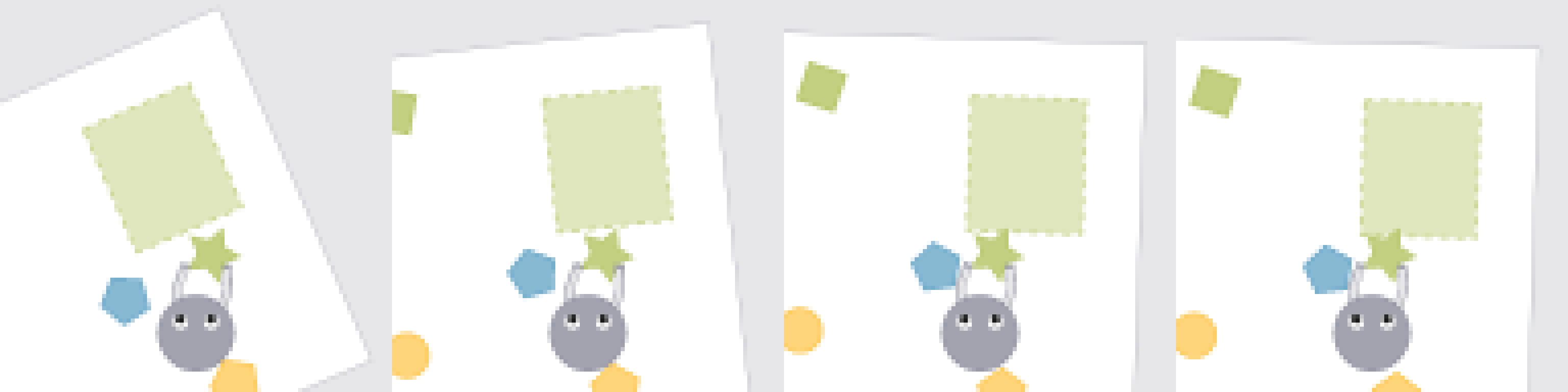}
        \caption{Egocentric}\label{fig:ego-allo-ego}
    \end{subfigure}
    \hspace{5mm}
    \begin{subfigure}[b]{0.45\textwidth}
        \centering
        \includegraphics[width=\textwidth]{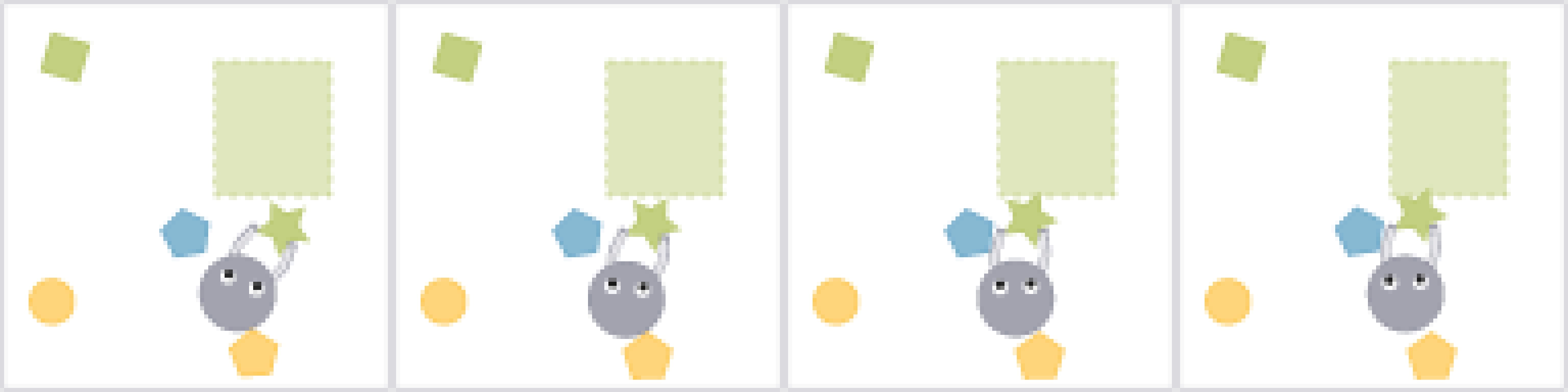}
        \caption{Allocentric}\label{fig:ego-allo-allo}
    \end{subfigure}
    \caption{
        Egocentric and allocentric views of a demonstration on MoveToRegion.
        The four 96$\times$96 RGB frames shown in each subfigure would normally be stacked together along the channels axis before being passed to an agent policy or discriminator.
    }
    \label{fig:ego-allo}
\end{figure}

We use the same discrete action space for all tasks.
Although this benchmark was inspired by robotic IL, where the underlying action space is generally continuous, we opted to use discrete actions so that we could elicit human demonstrations using only a standard keyboard.
The underlying state space is still continuous, so each discrete action applies a preset combination of forces to the robot, such as a force that pushes the gripper arms together, or a force that moves the robot forward or backward.
In total, the agent has 18 distinct actions.
These are formed from the Cartesian product of two gripper actions (push closed/allow to open), three longitudinal motion actions (forward/back/stop), and three angular motions (left/straight/right).

We use the same image-based observation space for each task.
In all of our experiments, we provide the agent with stacked 96$\times$96 pixel RGB frames depicting the workspace at the current time step and three preceding time steps.
At our 8Hz control rate, this corresponds to around 0.5s of interaction context.
Using an image-based observation space makes it easy to generalise policies and discriminators across different numbers and types of objects, without having to resort to, e.g., graph networks or structured learning.
An image-based observation space also means that the agent gets access to a similar representation as the human demonstrator.
This makes it possible to resolve ambiguities and improve generalisation by exploiting features of the human visual system, as we do when we apply the small image augmentations described in \cref{app:expt-details}.

By default, observations employ an egocentric (robot-centred) perspective on the workspace, as illustrated in \cref{fig:ego-allo-ego}.
Unlike the allocentric perspective, depicted in \cref{fig:ego-allo-allo}, the egocentric often does not allow the agent to observe the full workspace.
However, we found that an egocentric perspective resulted in faster training and better generalisation, as we note in the ablations of \cref{sec:expts}.
Similar benefits to generalisation were previously observed by \citet{hill2020environmental}.

\subsection{Detailed task descriptions}
    
\newcommand{\montw}{0.95\textwidth}  

\subsubsection{MoveToCorner (MTC)}

\begin{figure}[H]
    \centering
    \includegraphics[width=\montw]{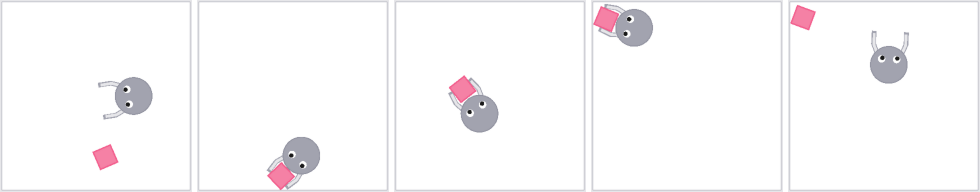}
    \caption{A demonstration on MoveToCorner.}
    \label{fig:demo-mtc}
\end{figure}

In MoveToCorner, the robot must push a single block from the bottom right corner of the workspace to the top left corner of the workspace.
Test variants are also constrained so that there is only ever one block, and it always starts close to the bottom right corner of the workspace.
These constraints preclude use of the CountPlus test variant, since block count cannot be changed without making the task ambiguous.
It also precludes use of the Layout variant, since fully randomising block position might make the desired block location ambiguous (e.g.\ pushing the block into top left corner versus pushing it to the opposite side of the workspace).
The horizon for all variants is $H = 80$ time steps.

Trajectories receive a score of $S(\tau)=1$ if the block spends the last frame of the rollout within $\sqrt{2}/2$ units of the top left corner of the workspace (the whole workspace is 2$\times$2 units).
$S(\tau)$ decays linearly from 1 to 0 as the block moves from inside that region to more than $\sqrt{2}$ units away from the corner.

\subsubsection{MoveToRegion (MTR)}

\begin{figure}[H]
    \centering
    \includegraphics[width=\montw]{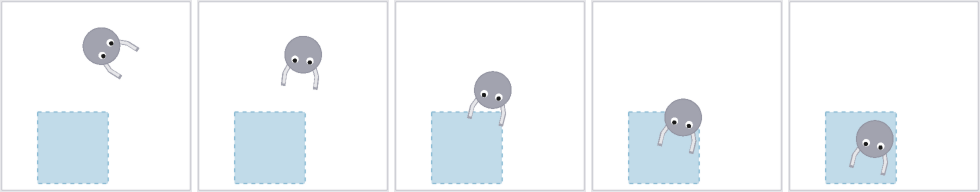}
    \caption{A demonstration on MoveToRegion.}
    \label{fig:demo-mtr}
\end{figure}

The objective of the MoveToRegion task is for the robot to drive inside a goal region placed in the workspace.
There is only ever one goal region, and no blocks are present in the train or test variants.
Hence the CountPlus and Shape variants are not applicable.
However, the Colour variant is still applicable. as it randomises the colour of the goal region.
The horizon is set to $H = 40$.

Scoring for MoveToRegion is binary.
If at the end of the episode, the centre of the robot's body is inside the goal region, then it receives a score of 1.
Otherwise it receives a score of 0.

\subsubsection{MatchRegions (MR)}

\begin{figure}[H]
    \centering
    \includegraphics[width=\montw]{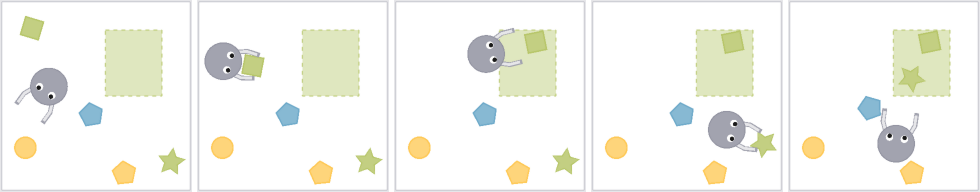}
    \caption{A demonstration on MatchRegions.}
    \label{fig:demo-mr}
\end{figure}

In MatchRegions, the agent is confronted with a single goal region and several blocks of different colours.
The objective is to move all (and only) blocks of the same colour as the goal region into the goal region.
All test variants are applicable to this version, although CountPlus only randomises the number of blocks (and not the number of goal regions) in order to avoid ambiguity about which goal region(s) the robot should fill with blocks.
The horizon is fixed to $H = 120$.

At the end of a trajectory $\tau$, the robot receives a score of
\[
  S(\tau) = \underbrace{\frac{|\mathcal T \cap \mathcal R|}{|\mathcal T|}}_{\text{Target bonus}} \times \underbrace{\left(1 - \frac{|\mathcal D \cap \mathcal R|}{|\mathcal R|}\right)}_{\text{Distractor penalty}}~.
\]
Here $\mathcal T$ is the set of \textit{target blocks} of the same colour as the goal region, $\mathcal D$ is the set of \textit{distractor blocks} of a different colour, and $\mathcal R$ is the set of blocks inside the goal region in the last state $s_T$ of the rollout $\tau$.
The agent gets a perfect score of 1 for placing all the target blocks and none of the distractors in the goal region.
Its score decreases for each target block it fails to move to the goal region (target bonus) and each distractor block it improperly places in the goal region (distractor penalty).

\subsubsection{MakeLine (ML)}

\begin{figure}[H]
    \centering
    \includegraphics[width=\montw]{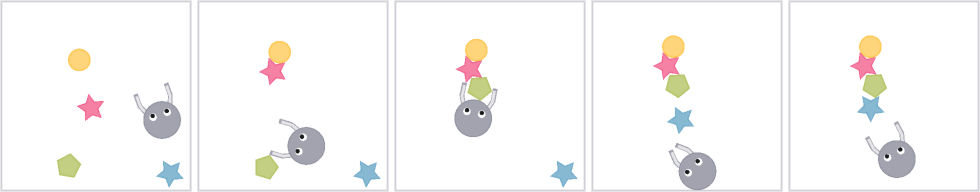}
    \caption{A demonstration on MakeLine.}
    \label{fig:demo-ml}
\end{figure}

The objective of the MakeLine task is to arrange all of the blocks in the workspace into a line.
The orientation and location of the line are ignored, as are the shapes and colours of the blocks involved.
The horizon for this task is $H = 180$.

Scoring for MakeLine is a function of the relative positions of blocks in the final state of a trajectory, and in particular the number of blocks that form the largest identifiable ``line''.
To identify lines of blocks, we use a line-fitting methods that is similar in spirit to RANSAC~\cite{bolles1981ransac}, but with constraints to ensure that blocks are spread out along the length of the line rather than ``bunching up''.
Our definition of what constitutes a line is based on a relation between triples of blocks: we say that a block $b_k$ is considered to be part of a line between blocks $b_i$ and $b_j$ if:
\begin{enumerate}
    \item \textbf{$b_k$ is an inlier:} it must lie a distance of at most $d_i = 0.18$ units from the (geometric) line that links $b_i$ and $b_j$ (recall that the workspace is $2 \times 2$ units).
    
    \item \textbf{$b_k$ is close to other blocks in the line:} if $b_k$ is not the first or last block in the line of blocks, then it must be a distance of at most $d_c = 0.42$ units from the previous and next blocks.
    Here the distance is measured along the direction of the geometric line between $b_i$ and $b_j$.
    That is, by projecting the previous and next inliers onto the geometric line between $b_i$ and $b_j$, then taking the distance between those projections and the projection for $b_k$.
\end{enumerate}
Note that if $b_i$ and $b_j$ are a long way apart, then there may be several subsets of inliers for the line between $b_i$ and $b_j$, each of which is separated from the other subsets than $d_c$ units.
For any given pair of blocks $(b_i, b_j)$, let $\#(b_i, b_j)$ be the number of blocks that form the \textit{largest} such subset for the line between $b_i$ and $b_j$ (potentially including $b_i$ and/or $b_j$, if they are close enough to the other inliers).
Further, let $n$ be the number of blocks in the workspace, and $m = \max_{i,j} \#(b_i, b_j)$ be the largest number of blocks on a line between any two blocks in the final state.
If $m = n$, then all blocks belong to the same line, and so $S(\tau) = 1$.
If $m = n - 1$, then exactly one block is not a part of the largest identifiable line, and $S(\tau) = 0.5$.
Otherwise, if $m < n - 1$, the agent receives a score of $S(\tau) = 0$.

\subsubsection{FindDupe (FD)}

\begin{figure}[H]
    \centering
    \includegraphics[width=\montw]{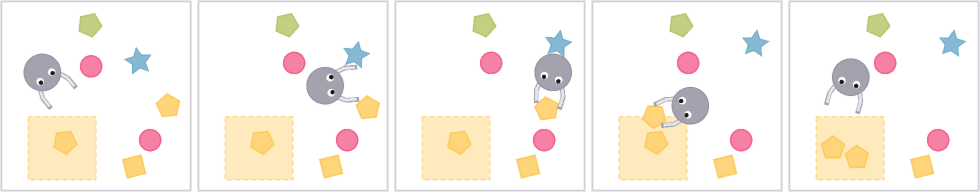}
    \caption{A demonstration on FindDupe.}
    \label{fig:demo-fd}
\end{figure}

FindDupe presents the agent with a goal region that has a single ``query'' block inside it, along with a mixture of blocks outside the goal region.
The agent's objective is to locate at least one block outside the goal region with the same shape and colour as the query block, and push it inside the goal region.
Variants are constrained so that there is only ever one goal region and query block, and so that there is at least one duplicate of the query block outside the goal region.
The horizon for this task is $H = 100$.

The score for this task is a function of the set of blocks present in the goal area at the end of the trajectory.
Let $\mathcal R$ denote the set of blocks inside the region at the end of the episode, let $\mathcal T$ denote the set of all \textit{target} blocks with the same shape and colour as the query block, and let $\mathcal D$ denote the set of all \textit{distractor} blocks with a different shape or colour.
Further, let $q$ refer to the original query block.
The score $S(\tau)$ for a trajectory is
\[
  S(\tau) = \underbrace{\mathbb I[q \in \mathcal R] \times \mathbb I[\mathcal T \cap \mathcal R \neq \varnothing]}_{\text{Query satisfied?}} \times \underbrace{\left(1 - \frac{|\mathcal D \cap \mathcal R|}{|\mathcal R|}\right)}_{\text{Distractor penalty}}~.
\]
The first factor ensures that the query block remains inside the goal region.
The second factor ensures that at least one other block with the same attributes as the query block is in the goal region.
Finally, the last factor creates a penalty for pushing distractor blocks into the goal region.

\subsubsection{FixColour (FC)}

\begin{figure}[H]
    \centering
    \includegraphics[width=\montw]{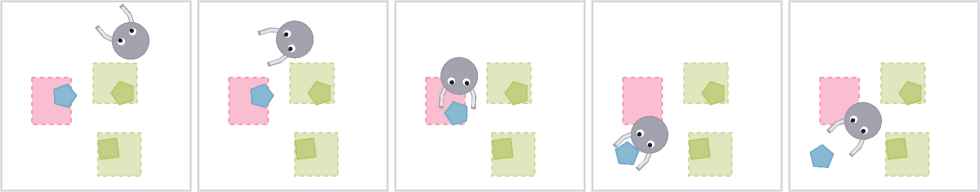}
    \caption{A demonstration on FixColour.}
    \label{fig:demo-fc}
\end{figure}

FixColour variants always include several non-overlapping goal regions, each containing a single block.
Exactly one of those blocks will be of a different colour to its enclosing goal region; we'll call this the ``mismatched block''.
The agent's objective is to identify the mismatched block and push it out of its goal region, into an unoccupied part of the workspace, thereby ``fixing'' the mismatch.
The horizon for this task is $H = 60$.

Scoring for FixColour is binary.
A score of $S(\tau) = 1$ is given if, in the final state, the mismatched block is not in its original goal region.
All other goal regions must contain exactly the same block that they started with (and in particular cannot contain the mismatched block).
If any of these conditions is not satisfied, then the score is zero.

\subsubsection{ClusterColour (CC) and ClusterShape (CS)}

\begin{figure}[H]
    \centering
    \includegraphics[width=\montw]{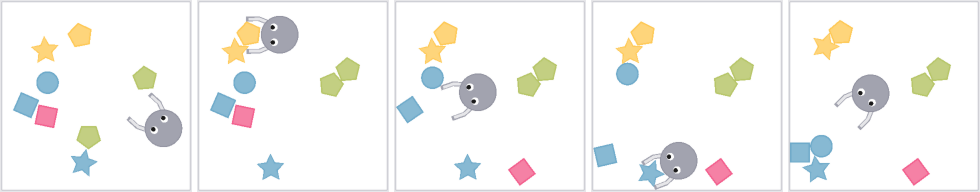}\\
    \vspace{0.5cm}
    \includegraphics[width=\montw]{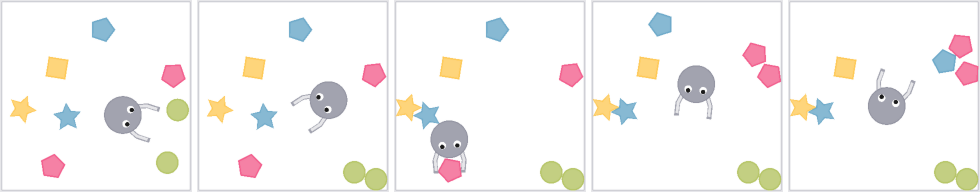}
    \caption{Demonstrations on ClusterColour (top) and ClusterShape (bottom).}
    \label{fig:demo-cc-cs}
\end{figure}

In both ClusterColour and ClusterShape, the workspace is initially filled with a jumble of blocks of different colours and types, and the agent must push the blocks into clusters according to some attribute.
For ClusterColour, blocks should belong to the same cluster iff they have the same colour, while ClusterShape applies the analogous criterion to block shape.
All variants are applicable to these tasks.
Because these tasks require interaction with most or all blocks in the workspace, the horizon is set to $H = 320$ (40s at 8Hz).

The score $S(\tau)$ takes the same form for both ClusterColour and ClusterShape, but with a different attribute-of-interest (either colour or shape).
Specifically, $S(\tau)$ is computed by applying a K-means-like objective to the final state $s_T$ of the rollout $\tau$.
For each value $a$ of the attribute-of-interest (either red/green/blue/yellow for ClusterColour or square/circle/pentagon/star for ClusterShape), a \textit{centroid} $x_a$ is computed from the mean positions of blocks with the corresponding attribute value.
Formally, this is
\[
    x_a = \frac{1}{|\mathcal B_a|} \sum_{b \in \mathcal B_a} b\text{.pos}~,
\]
where $\mathcal B_a$ is the set of blocks with the relevant attribute set to value $a$, and $b\text{.pos}$ is the position of block $b$ in state $s_T$.
In order for an individual block $b$ with relevant attribute value $a$ to be considered correctly clustered, the squared distance
\[
    d(b, a) = \| b.\text{pos} - x_a \|^2_2
\]
between it and its associated centroid must be at most a third the squared distance $d(b,a')$ between it and the nearest centroid for any other attribute value $a'$.
Specifically, we must have
\[
    d(b, a) < \frac{1}{3} \min_{a' \neq a} d(b,a')~.
\]
When 50\% or fewer of blocks are correctly clustered in the final state of a trajectory, the score $S(\tau) = 0$.
As the fraction of correctly clustered blocks increases from 50\% up to 100\%, the score $S(\tau)$ increases linearly from 0 to 1.

\section{Addition experiment details}\label{app:expt-details}

This section documents the full set of hyperparameters we used for BC and GAIL, along with additional details on how we collected and preprocessed our demonstrations.

\paragraph{Dataset and data preprocessing details}
%
We collected training datasets of 25 demonstration trajectories for the demonstration variant of each task.
These trajectories were recorded by the authors to show several distinct strategies for solving the task within the demonstration variant.
For instance, in ClusterColour, there are demonstrations that place clusters in different locations or construct them in a different order.
\cref{app:benchmark} shows a single demonstration for each task in the dataset.

Each algorithm run used only 10 of the 25 total trajectories for each task (or 10 trajectories for each task, in the multi-task case).
The subset of 10 trajectories was sampled at random based on the seed for that run.
We did not hold out any trajectories for testing or validation; rather, our evaluation is based on the test variant scores assigned to the trained policy produced by each algorithm.
For all policies, value functions, and discriminators, we constructed an observation by concatenating four temporally adjacent RGB frames along the channels axis, scaling the pixel values into the $[0,1]$ range, and resizing the stacked frames to 96$\times$96 pixels.
For BC, we performed the additional preprocessing step of removing samples with noop actions from the demonstration dataset, as described below.

\paragraph{Evaluation details}
%
%
For single-task BC and GAIL, we do five training runs on each task with different random seeds.
After each run, we take the trained policy, use it to perform 100 rollouts on each test variant of the original task, and retain the mean scores from those 100 trajectories.
In tables, we report ``mean score $\pm$ standard deviation of score'', where the mean and standard deviation are taken over the mean evaluation scores for each of the five runs on each algorithm and task.
Multitask evaluations are similar, except we pool data from all tasks together, and consequently only perform five runs in total rather than five runs per task.
To reduce variance, we used the same five random seeds (and consequently the same five subsets of 10 training trajectories each) for all algorithms and tasks.

\paragraph{Default augmentation set}
Throughout the text, we refer to noise, translation, rotation, and colour jitter augmentations.
Concretely, these augmentations involved the following operations:
\begin{itemize}
    \item \textbf{Noise:} Each (RGB) channel of each pixel is independently perturbed by additive noise sampled from $\mathcal N(0, 0.01)$.
    \item \textbf{Translation:} The image is mirror-padded and randomly translated along the $x$ and $y$ axes by up to 5\% of their respective range (so $\pm4.8$px, for 96$\times$96 pixels).
    \item \textbf{Rotation:} Image is mirror padded and then rotated around its centre by up to $\pm$5 degrees.
    \item \textbf{Colour jitter:} For this augmentation, images are translated to the CIELab colour space.
    The luminance channel is rescaled by a randomly sampled factor between 0.99 and 1.01, while the \textit{a} and \textit{b} channels are treated as a 2D vectors and randomly rotated by up to $\pm$0.15 radians.
    We use the same luminance scaling factor and colour rotation for each pixel an a given image.
    After these operations, images are converted back to RGB.
\end{itemize}
For the translation, rotation, and colour jitter augmentations, we apply the same randomly sampled transformation to each image in a four-image ``stack'' of frames, but different, independently sampled transformations to each stack in a training batch.

\paragraph{Single- and multi-task BC hyperparameters}
The hyperparameters for BC are given in \cref{tab:hp-bc}.
BC hyperparameters were manually tuned to ensure that losses plateaued on most single-task problems.
Note that hyperparameters for single- and multi-task learning were identical.
In particular, we retained the same batch size for multi-task experiments, and randomly sampled demonstration states from each task with a weighting that ensured equal representation from all tasks.
Initially, we found that training BC to convergence would cause the policy to get ``stuck'' in states where the most probable demonstrator action was a noop action.
We avoided this problem by removing all state/action pairs with noop actions from the dataset in our BC experiments; we did not do this in our GAIL experiments.

\begin{table}[t]
    \centering
    \begin{tabular}{lcc}
        \toprule
        \textbf{Hyperparameter} & \textbf{Value} & \textbf{Range Considered}\\\midrule
        Total opt. batches & 20,000 & 5,000--20,000\\
        Batch size & 32 & -\\
        SGD learning rate & $10^{-3}$ & -\\
        SGD momentum & 0.1 & -\\
        Policy augmentations & \makecell{Noise, trans.,\\ rot., colour jit.} & -\\
        \bottomrule
    \end{tabular}
    \vspace{2mm}
    \caption{Hyperarameters for BC experiments.}
    \label{tab:hp-bc}
\end{table}

\paragraph{Single- and multi-task GAIL hyperparameters}
Hyperparameters for GAIL are listed in \cref{tab:hp-gail}.
For policy optimisation, we used the PPO implementation from rlpyt~\cite{stooke2019rlpyt}; PPO hyperparameters that are not listed in \cref{tab:hp-gail} took their default values in rlpyt.
To prevent value and advantage magnitudes from exploding in PPO, we normalised rewards produced by the discriminator to have zero mean and a standard deviation of 0.1, both enforced using a running average and variance updated over the course of training.
Again, multi-task hyperparameters were the same as single-task hyperparameters, and we split each policy and discriminator training batch evenly between the tasks.

\begin{table}[t]
    \centering
    \begin{tabular}{lcc}
        \toprule
        \textbf{Hyperparameter} & \textbf{Value} & \textbf{Range Considered}\\\midrule
        Policy (PPO) & & \\
        \quad{}Sampler batch size & 32 & 16 to 64\\
        \quad{}Sampler time steps & 8 & 8 to 20\\
        \quad{}Opt. epochs per update & 12 & 2 to 10\\
        \quad{}Opt. minibatch size & 64 & 42 to 64\\
        \quad{}Initial Adam step size & $6\times10^{-5}$ & $10^{-6}$ to $10^{-3}$\\
        \quad{}Final Adam step size & 0 (lin.\ anneal) & -\\
        \quad{}Discount $\gamma$ & 0.8 & 0.8 to 1.0\\
        \quad{}GAE $\lambda$ & 0.8 & 0.8 to 1.0\\
        \quad{}Entropy bonus & $10^{-5}$ & $10^{-6}$ to $10^{-4}$\\
        \quad{}Advantage clip $\epsilon$ & 0.01 & 0.01 to 0.2\\
        \quad{}Grad. clip $\ell_2$ norm & 1.0 & -\\
        \quad{}Augmentations & N/A & -\\
        Discriminator & & \\
        \quad{}Batch size & 24 & -\\
        \quad{}Adam step size & $2.5\times 10^{-5}$ & $10^{-5}$ to $5\times 10^{-4}$\\
        \quad{}Augmentations & \makecell{Noise, trans.,\\ rot., colour jit.} & -\\
        \quad{}$\lambda_{\text{w-gp}}$ (WGAIL-GP) & 100 & -\\
        Misc. & &\\
        \quad{}Disc. steps per PPO update & 12 & 8 to 32\\
        \quad{}Total env. steps of training & $10^6$ & $5\times 10^5$ to $5 \times 10^6$\\
        \quad{}Reward norm. std. dev. & 0.1 & - \\
        \bottomrule
    \end{tabular}
    \vspace{2mm}
    \caption{Hyperarameters for GAIL experiments.}
    \label{tab:hp-gail}
\end{table}

\paragraph{Apprenticeship learning baseline}
In addition to our BC and (W)GAIL baselines, we also attempted to train a feature expectation matching Apprenticeship Learning (AL) baseline~\cite{abbeel2004apprenticeship,ho2016model}.
Given a feature function $\Phi : \mathcal S \times \mathcal A \to \mathbb R^n$, the goal of AL is to find a policy $\pi_\theta$ that matches the expected value of the feature function $\Phi$ under the demonstration distribution with its expected value under the novice distribution.
That is, we seek a $\pi_\theta$ such that $\expect_{\pi_\theta} \Phi(s,a) = \expect_{\mathcal D} \Phi(s,a)$.
Matching feature expectations is equivalent to finding a policy $\pi_\theta$ that drives the cost
\begin{equation}\label{eqn:al-sup}
    \sup_{\|w\|\leq 2} \left[
    \expect_{\mathcal D} w^T \Phi(s,a) - \expect_{\pi_\theta} w^T \Phi(s,a)  \right]
\end{equation}
to zero.
Observe that if $w^*$ is a weight vector that attains the supremum in \cref{eqn:al-sup}, then
\begin{align*}
    -\nabla_\theta \expect_{\pi_\theta} {w^*}^T \Phi(s,a)
\end{align*}
is a subgradient of \cref{eqn:al-sup} with respect to the policy parameters $\theta$.
Thus, our training procedure consisted of alternating between optimising \cref{eqn:al-sup} to convergence with respect to $w$, and taking a PPO step on the policy parameters using the reward function $r(s,a) = {w^*}^T \Phi(s,a)$ (recall that RL maximises return, but we want to minimise \cref{eqn:al-sup}).
To optimise $w$, we used 512 samples from the expert and the novice, and to optimise $\pi_\theta$, we used the same generator hyperparameters as our GAIL runs.
This single-task AL baseline is denoted ``AL (ST)'' in results tables.

The feature function $\Phi$ used for AL was acquired by removing the final (logit) layer of our GAIL discriminator network architecture and optimsing the remaining layers to minimise an autoencoder loss.
In creating the encoder, our only modification to the GAIL discriminator network architecture was to replace the 256-dimensional penultimate layer with a 32-dimensional one, to produce a 32-dimensional feature function $\Phi$.
This optimisation was performed for 8,192 size-24 batches of expert data, which we empirically found was enough to get clear reproduction of most input images.
After autoencoder pretraining, the encoder weights were kept frozen for the remainder of each training run.

Unfortunately, we could not get AL to produce adequate policies for any task except MoveToCorner.
We suspect that the poor performance of AL was due to inadequate autoencoder features.
The autoencoder was only trained on expert samples, and we found that for some problems it would not correctly reproduce images of states that were far from the support of the demonstrations.
It may be possible to improve results by training the autoencoder on both random rollouts and expert samples, or by training it on more diverse multi-task data.

\begin{figure}
    \centering
    \includegraphics{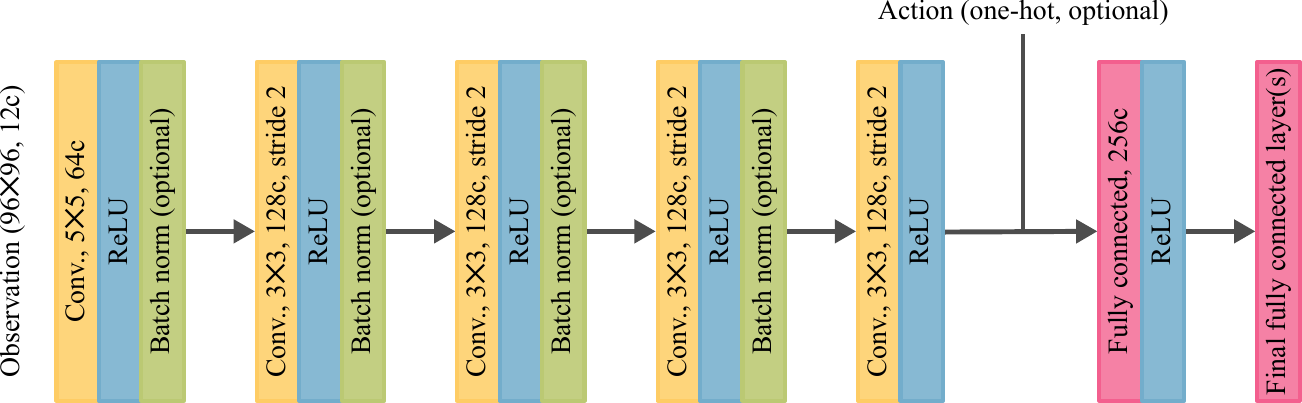}
    \caption{Base architecture for policies, value functions and discriminators.
    ``$n$c'' is used as an abbreviation for ``$n$ channels''.
    Refer to main text for a discussion of which networks use the optional features (batch norm, action input), and for a description of the final layer for each network type.}
    \label{fig:arch-diagram}
\end{figure}

\paragraph{Network architecture}
\cref{fig:arch-diagram} shows the base architecture for all neural networks used in the experiments (including discriminators, policies, and value functions).
Some experiments use slight variations on this basic policy architecture for some of the networks:
\begin{itemize}
    \item The one-hot action input is only used for discriminators, which concatenate the one-hot action representation to the activations of the final convolution layer before performing a forward pass through the linear layers.
    
    \item Batch norm is only used for the BC policy and GAIL discriminator, not for the GAIL policy and value function.

    \item In GAIL experiments, which train a policy via RL, the policy and value function share all layers \textit{except} the final fully-connected layer.

    \item In multitask experiments, the policy, value function, and discriminator share weights between tasks for all layers except the last.
    The final layer uses a single, separate set of weights corresponding to each task.
\end{itemize}

\paragraph{Computing infrastructure and experiment running time}
Experiments were performed on machines with 2$\times$ Xeon Gold 6130 CPUs (16 cores each, 2.1GHz base clock), 128--256GB RAM, and 4$\times$ GTX 1080-Ti GPUs.
Each ``run''---that is, the training and evaluation of a specific algorithm on a specific task with one seed---took an average of 10h03m (GAIL) and 32m (BC).
It should be noted that these wall time figures were recorded while performing up to 16 runs in parallel on each machine.
Because we did not use task-specific training durations, there was little variance in execution time between the different configurations (multi-task, egocentric, allocentric, etc.) of each of the two main base algorithms (BC and GAIL).
\section{Full experiment results}\label{app:results}

Full results for all methods, along with corresponding ablations, are shown in \cref{tab:results-tab1}, \cref{tab:results-tab2}, \cref{tab:results-tab3} and \cref{tab:results-tab4}.
We abbreviate behavioural cloning as ``BC'' and generative adversarial IL as ``GAIL'', while apprenticeship learning is ``AL''.
Single-task methods are denoted with ``(ST)'' and multi-task methods with ``(MT)''.
``Allo.'' is for experiments using an allocentric view; all other expeirments use an egocentric view.
For GAIL, ``WGAIL-GP'' denotes a version of GAIL that approximately minimises Wasserstein divergence while using a gradient penalty to encourage 1-Lipschitzness of the discriminator.
For augmentation ablations, we use ``no trans./rot. aug.'' to denote removal of translation/rotation; and ``no aug.'' to denote removal of all three default augmentations (colour, translation/rotation, Gaussian noise).

\begin{landscape}
    \begin{table}
        \centering
        \begin{footnotesize}
            \begin{tabular}{lcccccccc}
\toprule
     & \multicolumn{8}{c}{\textbf{MoveToCorner}} \\
        &                                           Demo &                                         Jitter & Layout &                                         Colour &                                          Shape & CountPlus &                                       Dynamics &                                            All \\
\midrule
AL (ST)                      &  \cellcolor[rgb]{0.99,1.00,1.00} 0.00$\pm$0.00 &  \cellcolor[rgb]{0.99,1.00,1.00} 0.00$\pm$0.00 &      - &  \cellcolor[rgb]{0.99,1.00,1.00} 0.00$\pm$0.00 &  \cellcolor[rgb]{0.99,1.00,1.00} 0.00$\pm$0.00 &         - &  \cellcolor[rgb]{0.99,1.00,1.00} 0.00$\pm$0.00 &  \cellcolor[rgb]{0.99,1.00,1.00} 0.00$\pm$0.00 \\
BC (MT)                      &  \cellcolor[rgb]{0.72,0.83,0.98} 0.97$\pm$0.04 &  \cellcolor[rgb]{0.73,0.85,0.98} 0.91$\pm$0.02 &      - &  \cellcolor[rgb]{0.81,0.89,0.96} 0.73$\pm$0.16 &  \cellcolor[rgb]{0.72,0.83,0.98} 0.98$\pm$0.01 &         - &  \cellcolor[rgb]{0.73,0.85,0.98} 0.92$\pm$0.04 &  \cellcolor[rgb]{0.85,0.91,0.96} 0.62$\pm$0.11 \\
BC (ST)                      &  \cellcolor[rgb]{0.72,0.83,0.98} 0.98$\pm$0.04 &  \cellcolor[rgb]{0.75,0.87,0.98} 0.86$\pm$0.07 &      - &  \cellcolor[rgb]{0.72,0.84,0.98} 0.96$\pm$0.05 &  \cellcolor[rgb]{0.72,0.83,0.98} 0.97$\pm$0.03 &         - &  \cellcolor[rgb]{0.73,0.85,0.98} 0.91$\pm$0.05 &  \cellcolor[rgb]{0.76,0.87,0.98} 0.84$\pm$0.06 \\
BC (ST, allo.)               &  \cellcolor[rgb]{0.72,0.84,0.98} 0.94$\pm$0.05 &  \cellcolor[rgb]{0.74,0.86,0.99} 0.89$\pm$0.04 &      - &  \cellcolor[rgb]{0.73,0.85,0.98} 0.93$\pm$0.04 &  \cellcolor[rgb]{0.72,0.83,0.98} 0.97$\pm$0.02 &         - &  \cellcolor[rgb]{0.73,0.86,0.99} 0.90$\pm$0.04 &  \cellcolor[rgb]{0.73,0.85,0.98} 0.91$\pm$0.02 \\
BC (ST, no aug.)             &  \cellcolor[rgb]{0.72,0.84,0.98} 0.96$\pm$0.04 &  \cellcolor[rgb]{0.80,0.89,0.97} 0.77$\pm$0.09 &      - &  \cellcolor[rgb]{0.78,0.88,0.97} 0.80$\pm$0.06 &  \cellcolor[rgb]{0.78,0.88,0.97} 0.81$\pm$0.12 &         - &  \cellcolor[rgb]{0.75,0.87,0.98} 0.86$\pm$0.05 &  \cellcolor[rgb]{0.86,0.92,0.96} 0.60$\pm$0.05 \\
BC (ST, no trans./rot. aug.) &  \cellcolor[rgb]{0.72,0.83,0.98} 0.96$\pm$0.04 &  \cellcolor[rgb]{0.76,0.87,0.98} 0.85$\pm$0.04 &      - &  \cellcolor[rgb]{0.76,0.87,0.98} 0.83$\pm$0.14 &  \cellcolor[rgb]{0.74,0.86,0.99} 0.88$\pm$0.06 &         - &  \cellcolor[rgb]{0.73,0.84,0.98} 0.94$\pm$0.05 &  \cellcolor[rgb]{0.84,0.91,0.95} 0.67$\pm$0.11 \\
GAIL (MT)                    &  \cellcolor[rgb]{0.94,0.97,0.98} 0.31$\pm$0.31 &  \cellcolor[rgb]{0.93,0.96,0.98} 0.33$\pm$0.31 &      - &  \cellcolor[rgb]{0.97,0.98,0.99} 0.16$\pm$0.09 &  \cellcolor[rgb]{0.93,0.96,0.98} 0.34$\pm$0.30 &         - &  \cellcolor[rgb]{0.94,0.97,0.98} 0.30$\pm$0.27 &  \cellcolor[rgb]{0.97,0.98,0.99} 0.16$\pm$0.10 \\
GAIL (ST)                    &  \cellcolor[rgb]{0.71,0.82,0.98} 0.99$\pm$0.01 &  \cellcolor[rgb]{0.73,0.85,0.99} 0.91$\pm$0.06 &      - &  \cellcolor[rgb]{0.79,0.88,0.97} 0.78$\pm$0.10 &  \cellcolor[rgb]{0.72,0.84,0.98} 0.95$\pm$0.03 &         - &  \cellcolor[rgb]{0.72,0.84,0.98} 0.95$\pm$0.05 &  \cellcolor[rgb]{0.85,0.91,0.95} 0.65$\pm$0.16 \\
GAIL (ST, allo.)             &  \cellcolor[rgb]{0.71,0.82,0.98} 1.00$\pm$0.00 &  \cellcolor[rgb]{0.77,0.88,0.97} 0.82$\pm$0.05 &      - &  \cellcolor[rgb]{0.73,0.85,0.99} 0.90$\pm$0.08 &  \cellcolor[rgb]{0.71,0.82,0.98} 0.99$\pm$0.01 &         - &  \cellcolor[rgb]{0.71,0.82,0.98} 0.99$\pm$0.01 &  \cellcolor[rgb]{0.86,0.92,0.96} 0.59$\pm$0.12 \\
GAIL (ST, no aug.)           &  \cellcolor[rgb]{0.87,0.93,0.96} 0.56$\pm$0.36 &  \cellcolor[rgb]{0.93,0.96,0.98} 0.36$\pm$0.23 &      - &  \cellcolor[rgb]{0.92,0.96,0.98} 0.39$\pm$0.30 &  \cellcolor[rgb]{0.93,0.96,0.98} 0.34$\pm$0.24 &         - &  \cellcolor[rgb]{0.90,0.94,0.97} 0.46$\pm$0.27 &  \cellcolor[rgb]{0.98,0.99,0.99} 0.11$\pm$0.10 \\
WGAIL-GP (ST)                &  \cellcolor[rgb]{0.93,0.96,0.98} 0.35$\pm$0.24 &  \cellcolor[rgb]{0.96,0.97,0.99} 0.22$\pm$0.12 &      - &  \cellcolor[rgb]{0.97,0.98,0.99} 0.17$\pm$0.20 &  \cellcolor[rgb]{0.94,0.96,0.98} 0.32$\pm$0.21 &         - &  \cellcolor[rgb]{0.94,0.97,0.98} 0.30$\pm$0.21 &  \cellcolor[rgb]{0.99,0.99,1.00} 0.04$\pm$0.05 \\
\midrule
     & \multicolumn{8}{c}{\textbf{MoveToRegion}} \\
        &                                           Demo &                                         Jitter &                                         Layout &                                         Colour & Shape & CountPlus &                                       Dynamics &                                            All \\
\midrule
AL (ST)                      &  \cellcolor[rgb]{0.88,0.93,0.97} 0.51$\pm$0.42 &  \cellcolor[rgb]{0.89,0.94,0.97} 0.47$\pm$0.39 &  \cellcolor[rgb]{0.96,0.98,0.99} 0.22$\pm$0.17 &  \cellcolor[rgb]{0.96,0.98,0.99} 0.21$\pm$0.22 &     - &         - &  \cellcolor[rgb]{0.88,0.94,0.97} 0.51$\pm$0.41 &  \cellcolor[rgb]{0.98,0.99,1.00} 0.09$\pm$0.05 \\
BC (MT)                      &  \cellcolor[rgb]{0.79,0.88,0.97} 0.79$\pm$0.22 &  \cellcolor[rgb]{0.79,0.89,0.97} 0.77$\pm$0.26 &  \cellcolor[rgb]{0.88,0.93,0.97} 0.52$\pm$0.13 &  \cellcolor[rgb]{0.86,0.92,0.96} 0.60$\pm$0.16 &     - &         - &  \cellcolor[rgb]{0.78,0.88,0.97} 0.81$\pm$0.24 &  \cellcolor[rgb]{0.95,0.97,0.99} 0.26$\pm$0.04 \\
BC (ST)                      &  \cellcolor[rgb]{0.74,0.86,0.99} 0.89$\pm$0.11 &  \cellcolor[rgb]{0.74,0.86,0.99} 0.88$\pm$0.11 &  \cellcolor[rgb]{0.89,0.94,0.97} 0.48$\pm$0.12 &  \cellcolor[rgb]{0.86,0.92,0.96} 0.60$\pm$0.13 &     - &         - &  \cellcolor[rgb]{0.74,0.86,0.99} 0.88$\pm$0.10 &  \cellcolor[rgb]{0.95,0.97,0.98} 0.28$\pm$0.06 \\
BC (ST, allo.)               &  \cellcolor[rgb]{0.85,0.91,0.95} 0.63$\pm$0.08 &  \cellcolor[rgb]{0.87,0.92,0.96} 0.57$\pm$0.14 &  \cellcolor[rgb]{0.98,0.99,1.00} 0.09$\pm$0.02 &  \cellcolor[rgb]{0.87,0.93,0.96} 0.56$\pm$0.19 &     - &         - &  \cellcolor[rgb]{0.86,0.92,0.96} 0.61$\pm$0.12 &  \cellcolor[rgb]{0.98,0.99,1.00} 0.10$\pm$0.02 \\
BC (ST, no aug.)             &  \cellcolor[rgb]{0.74,0.86,0.99} 0.88$\pm$0.12 &  \cellcolor[rgb]{0.76,0.87,0.98} 0.83$\pm$0.11 &  \cellcolor[rgb]{0.90,0.95,0.97} 0.44$\pm$0.08 &  \cellcolor[rgb]{0.80,0.89,0.96} 0.75$\pm$0.17 &     - &         - &  \cellcolor[rgb]{0.74,0.86,0.99} 0.87$\pm$0.12 &  \cellcolor[rgb]{0.93,0.96,0.98} 0.36$\pm$0.07 \\
BC (ST, no trans./rot. aug.) &  \cellcolor[rgb]{0.73,0.85,0.99} 0.91$\pm$0.05 &  \cellcolor[rgb]{0.76,0.87,0.98} 0.85$\pm$0.10 &  \cellcolor[rgb]{0.90,0.95,0.97} 0.44$\pm$0.10 &  \cellcolor[rgb]{0.81,0.89,0.96} 0.73$\pm$0.13 &     - &         - &  \cellcolor[rgb]{0.74,0.86,0.99} 0.89$\pm$0.06 &  \cellcolor[rgb]{0.94,0.96,0.98} 0.33$\pm$0.05 \\
GAIL (MT)                    &  \cellcolor[rgb]{0.71,0.82,0.98} 1.00$\pm$0.00 &  \cellcolor[rgb]{0.71,0.82,0.98} 0.99$\pm$0.02 &  \cellcolor[rgb]{0.83,0.90,0.96} 0.69$\pm$0.20 &  \cellcolor[rgb]{0.93,0.96,0.98} 0.34$\pm$0.09 &     - &         - &  \cellcolor[rgb]{0.71,0.82,0.98} 1.00$\pm$0.00 &  \cellcolor[rgb]{0.94,0.97,0.98} 0.31$\pm$0.07 \\
GAIL (ST)                    &  \cellcolor[rgb]{0.71,0.82,0.98} 1.00$\pm$0.00 &  \cellcolor[rgb]{0.71,0.82,0.98} 1.00$\pm$0.00 &  \cellcolor[rgb]{0.82,0.90,0.96} 0.71$\pm$0.09 &  \cellcolor[rgb]{0.92,0.95,0.98} 0.40$\pm$0.07 &     - &         - &  \cellcolor[rgb]{0.71,0.82,0.98} 1.00$\pm$0.00 &  \cellcolor[rgb]{0.95,0.97,0.98} 0.29$\pm$0.07 \\
GAIL (ST, allo.)             &  \cellcolor[rgb]{0.71,0.82,0.98} 1.00$\pm$0.00 &  \cellcolor[rgb]{0.72,0.83,0.98} 0.98$\pm$0.02 &  \cellcolor[rgb]{0.98,0.99,1.00} 0.08$\pm$0.02 &  \cellcolor[rgb]{0.72,0.84,0.98} 0.95$\pm$0.02 &     - &         - &  \cellcolor[rgb]{0.71,0.82,0.98} 1.00$\pm$0.00 &  \cellcolor[rgb]{0.98,0.99,1.00} 0.10$\pm$0.03 \\
GAIL (ST, no aug.)           &  \cellcolor[rgb]{0.71,0.82,0.98} 0.99$\pm$0.03 &  \cellcolor[rgb]{0.79,0.88,0.97} 0.79$\pm$0.08 &  \cellcolor[rgb]{0.93,0.96,0.98} 0.34$\pm$0.10 &  \cellcolor[rgb]{0.87,0.93,0.96} 0.56$\pm$0.12 &     - &         - &  \cellcolor[rgb]{0.73,0.85,0.98} 0.93$\pm$0.06 &  \cellcolor[rgb]{0.96,0.98,0.99} 0.20$\pm$0.06 \\
WGAIL-GP (ST)                &  \cellcolor[rgb]{0.72,0.84,0.98} 0.94$\pm$0.03 &  \cellcolor[rgb]{0.74,0.86,0.99} 0.87$\pm$0.04 &  \cellcolor[rgb]{0.86,0.92,0.96} 0.60$\pm$0.06 &  \cellcolor[rgb]{0.96,0.97,0.99} 0.23$\pm$0.01 &     - &         - &  \cellcolor[rgb]{0.73,0.84,0.98} 0.94$\pm$0.03 &  \cellcolor[rgb]{0.97,0.98,0.99} 0.18$\pm$0.04 \\
\bottomrule
\end{tabular}
        \end{footnotesize}
        \vspace{3mm}
        \caption{
            Scores for all compared methods on two tasks, reported as ``mean (std.)'' over five training runs (individual run means were computed with 100 rollouts each).
            A colour scale (\crule{0.99,1.00,1.00}\!\crule{0.97,0.98,0.99}\!\crule{0.95,0.97,0.98}\!\crule{0.91,0.95,0.97}\!\crule{0.87,0.92,0.96}\!\crule{0.82,0.90,0.96}\!\crule{0.75,0.87,0.98}\!\crule{0.71,0.82,0.98}) grades mean scores from poor (lightest) to perfect (darkest).
            See main text in \cref{app:results} for abbreviations.
        }\label{tab:results-tab1}
    \end{table}
\end{landscape}

\begin{landscape}
    \begin{table}
        \centering
        \begin{footnotesize}
            \begin{tabular}{lcccccccc}
\toprule
     & \multicolumn{8}{c}{\textbf{MatchRegions}} \\
        &                                           Demo &                                         Jitter &                                         Layout &                                         Colour &                                          Shape &                                      CountPlus &                                       Dynamics &                                            All \\
\midrule
AL (ST)                      &  \cellcolor[rgb]{0.99,1.00,1.00} 0.00$\pm$0.00 &  \cellcolor[rgb]{0.99,1.00,1.00} 0.00$\pm$0.00 &  \cellcolor[rgb]{0.99,1.00,1.00} 0.01$\pm$0.01 &  \cellcolor[rgb]{0.99,1.00,1.00} 0.00$\pm$0.00 &  \cellcolor[rgb]{0.99,1.00,1.00} 0.00$\pm$0.00 &  \cellcolor[rgb]{0.99,1.00,1.00} 0.01$\pm$0.01 &  \cellcolor[rgb]{0.99,1.00,1.00} 0.00$\pm$0.00 &  \cellcolor[rgb]{0.99,1.00,1.00} 0.01$\pm$0.01 \\
BC (MT)                      &  \cellcolor[rgb]{0.83,0.90,0.96} 0.69$\pm$0.10 &  \cellcolor[rgb]{0.86,0.92,0.96} 0.60$\pm$0.09 &  \cellcolor[rgb]{0.99,0.99,1.00} 0.05$\pm$0.01 &  \cellcolor[rgb]{0.94,0.96,0.98} 0.32$\pm$0.06 &  \cellcolor[rgb]{0.84,0.91,0.95} 0.65$\pm$0.07 &  \cellcolor[rgb]{0.99,0.99,1.00} 0.04$\pm$0.02 &  \cellcolor[rgb]{0.85,0.91,0.95} 0.64$\pm$0.09 &  \cellcolor[rgb]{0.99,0.99,1.00} 0.04$\pm$0.01 \\
BC (ST)                      &  \cellcolor[rgb]{0.80,0.89,0.97} 0.77$\pm$0.09 &  \cellcolor[rgb]{0.83,0.90,0.96} 0.69$\pm$0.11 &  \cellcolor[rgb]{0.98,0.99,1.00} 0.07$\pm$0.02 &  \cellcolor[rgb]{0.91,0.95,0.97} 0.42$\pm$0.04 &  \cellcolor[rgb]{0.83,0.90,0.96} 0.69$\pm$0.10 &  \cellcolor[rgb]{0.98,0.99,1.00} 0.07$\pm$0.03 &  \cellcolor[rgb]{0.83,0.90,0.96} 0.68$\pm$0.08 &  \cellcolor[rgb]{0.99,0.99,1.00} 0.05$\pm$0.02 \\
BC (ST, allo.)               &  \cellcolor[rgb]{0.82,0.90,0.96} 0.72$\pm$0.06 &  \cellcolor[rgb]{0.86,0.92,0.96} 0.58$\pm$0.14 &  \cellcolor[rgb]{0.99,1.00,1.00} 0.01$\pm$0.01 &  \cellcolor[rgb]{0.86,0.92,0.96} 0.58$\pm$0.11 &  \cellcolor[rgb]{0.86,0.92,0.96} 0.60$\pm$0.09 &  \cellcolor[rgb]{0.99,0.99,1.00} 0.02$\pm$0.01 &  \cellcolor[rgb]{0.87,0.93,0.96} 0.56$\pm$0.10 &  \cellcolor[rgb]{0.99,0.99,1.00} 0.03$\pm$0.02 \\
BC (ST, no aug.)             &  \cellcolor[rgb]{0.82,0.90,0.96} 0.71$\pm$0.05 &  \cellcolor[rgb]{0.89,0.94,0.97} 0.49$\pm$0.07 &  \cellcolor[rgb]{0.99,0.99,1.00} 0.04$\pm$0.01 &  \cellcolor[rgb]{0.93,0.96,0.98} 0.35$\pm$0.05 &  \cellcolor[rgb]{0.86,0.92,0.96} 0.59$\pm$0.05 &  \cellcolor[rgb]{0.99,0.99,1.00} 0.05$\pm$0.02 &  \cellcolor[rgb]{0.87,0.93,0.96} 0.54$\pm$0.05 &  \cellcolor[rgb]{0.99,0.99,1.00} 0.04$\pm$0.02 \\
BC (ST, no trans./rot. aug.) &  \cellcolor[rgb]{0.80,0.89,0.96} 0.75$\pm$0.05 &  \cellcolor[rgb]{0.87,0.93,0.96} 0.54$\pm$0.05 &  \cellcolor[rgb]{0.99,0.99,1.00} 0.06$\pm$0.01 &  \cellcolor[rgb]{0.93,0.96,0.98} 0.34$\pm$0.03 &  \cellcolor[rgb]{0.85,0.91,0.96} 0.62$\pm$0.07 &  \cellcolor[rgb]{0.98,0.99,1.00} 0.06$\pm$0.01 &  \cellcolor[rgb]{0.85,0.91,0.96} 0.63$\pm$0.02 &  \cellcolor[rgb]{0.99,0.99,1.00} 0.05$\pm$0.02 \\
GAIL (MT)                    &  \cellcolor[rgb]{0.97,0.98,0.99} 0.19$\pm$0.12 &  \cellcolor[rgb]{0.96,0.98,0.99} 0.20$\pm$0.10 &  \cellcolor[rgb]{0.99,0.99,1.00} 0.05$\pm$0.02 &  \cellcolor[rgb]{0.98,0.99,1.00} 0.07$\pm$0.03 &  \cellcolor[rgb]{0.96,0.98,0.99} 0.19$\pm$0.12 &  \cellcolor[rgb]{0.99,0.99,1.00} 0.02$\pm$0.01 &  \cellcolor[rgb]{0.96,0.98,0.99} 0.20$\pm$0.11 &  \cellcolor[rgb]{0.99,0.99,1.00} 0.04$\pm$0.01 \\
GAIL (ST)                    &  \cellcolor[rgb]{0.73,0.84,0.98} 0.94$\pm$0.03 &  \cellcolor[rgb]{0.73,0.85,0.98} 0.92$\pm$0.03 &  \cellcolor[rgb]{0.96,0.98,0.99} 0.21$\pm$0.02 &  \cellcolor[rgb]{0.94,0.97,0.98} 0.31$\pm$0.10 &  \cellcolor[rgb]{0.73,0.85,0.98} 0.93$\pm$0.05 &  \cellcolor[rgb]{0.97,0.98,0.99} 0.14$\pm$0.04 &  \cellcolor[rgb]{0.73,0.85,0.98} 0.92$\pm$0.04 &  \cellcolor[rgb]{0.97,0.98,0.99} 0.14$\pm$0.04 \\
GAIL (ST, allo.)             &  \cellcolor[rgb]{0.85,0.91,0.95} 0.64$\pm$0.13 &  \cellcolor[rgb]{0.86,0.92,0.96} 0.58$\pm$0.11 &  \cellcolor[rgb]{0.99,1.00,1.00} 0.01$\pm$0.01 &  \cellcolor[rgb]{0.93,0.96,0.98} 0.36$\pm$0.08 &  \cellcolor[rgb]{0.85,0.92,0.96} 0.62$\pm$0.11 &  \cellcolor[rgb]{0.99,1.00,1.00} 0.01$\pm$0.01 &  \cellcolor[rgb]{0.86,0.92,0.96} 0.57$\pm$0.11 &  \cellcolor[rgb]{0.99,1.00,1.00} 0.02$\pm$0.02 \\
GAIL (ST, no aug.)           &  \cellcolor[rgb]{0.90,0.95,0.97} 0.44$\pm$0.24 &  \cellcolor[rgb]{0.93,0.96,0.98} 0.35$\pm$0.18 &  \cellcolor[rgb]{0.99,0.99,1.00} 0.04$\pm$0.03 &  \cellcolor[rgb]{0.97,0.98,0.99} 0.18$\pm$0.13 &  \cellcolor[rgb]{0.93,0.96,0.98} 0.35$\pm$0.20 &  \cellcolor[rgb]{0.99,0.99,1.00} 0.03$\pm$0.02 &  \cellcolor[rgb]{0.93,0.96,0.98} 0.36$\pm$0.19 &  \cellcolor[rgb]{0.99,0.99,1.00} 0.02$\pm$0.01 \\
WGAIL-GP (ST)                &  \cellcolor[rgb]{0.94,0.96,0.98} 0.32$\pm$0.05 &  \cellcolor[rgb]{0.94,0.97,0.98} 0.30$\pm$0.04 &  \cellcolor[rgb]{0.97,0.98,0.99} 0.15$\pm$0.04 &  \cellcolor[rgb]{0.98,0.99,1.00} 0.07$\pm$0.00 &  \cellcolor[rgb]{0.94,0.96,0.98} 0.32$\pm$0.05 &  \cellcolor[rgb]{0.98,0.99,0.99} 0.11$\pm$0.03 &  \cellcolor[rgb]{0.95,0.97,0.99} 0.28$\pm$0.03 &  \cellcolor[rgb]{0.98,0.99,1.00} 0.08$\pm$0.02 \\
\midrule
     & \multicolumn{8}{c}{\textbf{MakeLine}} \\
        &                                           Demo &                                         Jitter &                                         Layout &                                         Colour &                                          Shape &                                      CountPlus &                                       Dynamics &                                            All \\
\midrule
AL (ST)                      &  \cellcolor[rgb]{0.99,1.00,1.00} 0.00$\pm$0.00 &  \cellcolor[rgb]{0.99,1.00,1.00} 0.00$\pm$0.00 &  \cellcolor[rgb]{0.99,0.99,1.00} 0.04$\pm$0.01 &  \cellcolor[rgb]{0.99,1.00,1.00} 0.00$\pm$0.00 &  \cellcolor[rgb]{0.99,1.00,1.00} 0.00$\pm$0.00 &  \cellcolor[rgb]{0.99,0.99,1.00} 0.03$\pm$0.01 &  \cellcolor[rgb]{0.99,1.00,1.00} 0.00$\pm$0.00 &  \cellcolor[rgb]{0.99,0.99,1.00} 0.03$\pm$0.01 \\
BC (MT)                      &  \cellcolor[rgb]{0.94,0.97,0.98} 0.31$\pm$0.07 &  \cellcolor[rgb]{0.95,0.97,0.98} 0.29$\pm$0.02 &  \cellcolor[rgb]{0.97,0.98,0.99} 0.18$\pm$0.03 &  \cellcolor[rgb]{0.97,0.98,0.99} 0.18$\pm$0.03 &  \cellcolor[rgb]{0.94,0.97,0.98} 0.30$\pm$0.05 &  \cellcolor[rgb]{0.97,0.98,0.99} 0.16$\pm$0.02 &  \cellcolor[rgb]{0.95,0.97,0.99} 0.28$\pm$0.07 &  \cellcolor[rgb]{0.97,0.98,0.99} 0.14$\pm$0.02 \\
BC (ST)                      &  \cellcolor[rgb]{0.89,0.94,0.97} 0.48$\pm$0.08 &  \cellcolor[rgb]{0.91,0.95,0.97} 0.43$\pm$0.07 &  \cellcolor[rgb]{0.96,0.98,0.99} 0.20$\pm$0.04 &  \cellcolor[rgb]{0.94,0.96,0.98} 0.32$\pm$0.04 &  \cellcolor[rgb]{0.91,0.95,0.97} 0.41$\pm$0.05 &  \cellcolor[rgb]{0.97,0.98,0.99} 0.18$\pm$0.05 &  \cellcolor[rgb]{0.91,0.95,0.97} 0.42$\pm$0.07 &  \cellcolor[rgb]{0.97,0.98,0.99} 0.18$\pm$0.04 \\
BC (ST, allo.)               &  \cellcolor[rgb]{0.96,0.97,0.99} 0.25$\pm$0.07 &  \cellcolor[rgb]{0.96,0.97,0.99} 0.24$\pm$0.04 &  \cellcolor[rgb]{0.99,0.99,1.00} 0.04$\pm$0.01 &  \cellcolor[rgb]{0.98,0.99,0.99} 0.11$\pm$0.02 &  \cellcolor[rgb]{0.96,0.98,0.99} 0.21$\pm$0.07 &  \cellcolor[rgb]{0.99,0.99,1.00} 0.03$\pm$0.01 &  \cellcolor[rgb]{0.96,0.98,0.99} 0.19$\pm$0.03 &  \cellcolor[rgb]{0.99,0.99,1.00} 0.02$\pm$0.01 \\
BC (ST, no aug.)             &  \cellcolor[rgb]{0.96,0.98,0.99} 0.19$\pm$0.03 &  \cellcolor[rgb]{0.98,0.99,0.99} 0.12$\pm$0.06 &  \cellcolor[rgb]{0.98,0.99,0.99} 0.11$\pm$0.03 &  \cellcolor[rgb]{0.98,0.99,0.99} 0.11$\pm$0.02 &  \cellcolor[rgb]{0.97,0.98,0.99} 0.14$\pm$0.03 &  \cellcolor[rgb]{0.98,0.99,1.00} 0.09$\pm$0.02 &  \cellcolor[rgb]{0.98,0.99,0.99} 0.11$\pm$0.05 &  \cellcolor[rgb]{0.98,0.99,1.00} 0.09$\pm$0.02 \\
BC (ST, no trans./rot. aug.) &  \cellcolor[rgb]{0.96,0.97,0.99} 0.23$\pm$0.05 &  \cellcolor[rgb]{0.97,0.98,0.99} 0.17$\pm$0.06 &  \cellcolor[rgb]{0.97,0.98,0.99} 0.12$\pm$0.04 &  \cellcolor[rgb]{0.97,0.98,0.99} 0.13$\pm$0.02 &  \cellcolor[rgb]{0.97,0.98,0.99} 0.15$\pm$0.02 &  \cellcolor[rgb]{0.98,0.99,0.99} 0.12$\pm$0.06 &  \cellcolor[rgb]{0.97,0.98,0.99} 0.14$\pm$0.04 &  \cellcolor[rgb]{0.98,0.99,0.99} 0.11$\pm$0.03 \\
GAIL (MT)                    &  \cellcolor[rgb]{0.99,1.00,1.00} 0.02$\pm$0.01 &  \cellcolor[rgb]{0.99,1.00,1.00} 0.01$\pm$0.01 &  \cellcolor[rgb]{0.99,0.99,1.00} 0.06$\pm$0.02 &  \cellcolor[rgb]{0.99,1.00,1.00} 0.02$\pm$0.00 &  \cellcolor[rgb]{0.99,1.00,1.00} 0.02$\pm$0.01 &  \cellcolor[rgb]{0.99,0.99,1.00} 0.05$\pm$0.02 &  \cellcolor[rgb]{0.99,1.00,1.00} 0.02$\pm$0.01 &  \cellcolor[rgb]{0.99,0.99,1.00} 0.06$\pm$0.03 \\
GAIL (ST)                    &  \cellcolor[rgb]{0.94,0.96,0.98} 0.33$\pm$0.19 &  \cellcolor[rgb]{0.94,0.96,0.98} 0.33$\pm$0.21 &  \cellcolor[rgb]{0.96,0.98,0.99} 0.20$\pm$0.02 &  \cellcolor[rgb]{0.97,0.98,0.99} 0.19$\pm$0.07 &  \cellcolor[rgb]{0.95,0.97,0.99} 0.27$\pm$0.16 &  \cellcolor[rgb]{0.97,0.98,0.99} 0.17$\pm$0.05 &  \cellcolor[rgb]{0.95,0.97,0.99} 0.28$\pm$0.15 &  \cellcolor[rgb]{0.97,0.98,0.99} 0.17$\pm$0.04 \\
GAIL (ST, allo.)             &  \cellcolor[rgb]{0.99,1.00,1.00} 0.01$\pm$0.01 &  \cellcolor[rgb]{0.99,1.00,1.00} 0.01$\pm$0.01 &  \cellcolor[rgb]{0.99,0.99,1.00} 0.05$\pm$0.02 &  \cellcolor[rgb]{0.99,1.00,1.00} 0.01$\pm$0.01 &  \cellcolor[rgb]{0.99,1.00,1.00} 0.01$\pm$0.01 &  \cellcolor[rgb]{0.99,0.99,1.00} 0.04$\pm$0.01 &  \cellcolor[rgb]{0.99,1.00,1.00} 0.01$\pm$0.01 &  \cellcolor[rgb]{0.99,0.99,1.00} 0.03$\pm$0.01 \\
GAIL (ST, no aug.)           &  \cellcolor[rgb]{0.99,0.99,1.00} 0.05$\pm$0.03 &  \cellcolor[rgb]{0.99,0.99,1.00} 0.03$\pm$0.02 &  \cellcolor[rgb]{0.99,0.99,1.00} 0.05$\pm$0.02 &  \cellcolor[rgb]{0.99,0.99,1.00} 0.02$\pm$0.01 &  \cellcolor[rgb]{0.99,0.99,1.00} 0.03$\pm$0.02 &  \cellcolor[rgb]{0.99,0.99,1.00} 0.05$\pm$0.02 &  \cellcolor[rgb]{0.99,0.99,1.00} 0.04$\pm$0.03 &  \cellcolor[rgb]{0.99,0.99,1.00} 0.04$\pm$0.02 \\
WGAIL-GP (ST)                &  \cellcolor[rgb]{0.98,0.99,1.00} 0.10$\pm$0.04 &  \cellcolor[rgb]{0.98,0.99,1.00} 0.08$\pm$0.03 &  \cellcolor[rgb]{0.98,0.99,0.99} 0.11$\pm$0.01 &  \cellcolor[rgb]{0.99,0.99,1.00} 0.06$\pm$0.02 &  \cellcolor[rgb]{0.98,0.99,1.00} 0.07$\pm$0.03 &  \cellcolor[rgb]{0.98,0.99,0.99} 0.11$\pm$0.04 &  \cellcolor[rgb]{0.98,0.99,1.00} 0.08$\pm$0.02 &  \cellcolor[rgb]{0.98,0.99,1.00} 0.08$\pm$0.01 \\
\bottomrule
\end{tabular}
        \end{footnotesize}
        \vspace{3mm}
        \caption{
            Additional results; refer to \cref{tab:results-tab1} for details.
        }\label{tab:results-tab2}
    \end{table}
\end{landscape}

\begin{landscape}
    \begin{table}
        \centering
        \begin{footnotesize}
            \begin{tabular}{lcccccccc}
\toprule
     & \multicolumn{8}{c}{\textbf{FindDupe}} \\
        &                                           Demo &                                         Jitter &                                         Layout &                                         Colour &                                          Shape &                                      CountPlus &                                       Dynamics &                                            All \\
\midrule
AL (ST)                      &  \cellcolor[rgb]{0.99,1.00,1.00} 0.00$\pm$0.00 &  \cellcolor[rgb]{0.99,1.00,1.00} 0.00$\pm$0.00 &  \cellcolor[rgb]{0.99,1.00,1.00} 0.00$\pm$0.00 &  \cellcolor[rgb]{0.99,1.00,1.00} 0.00$\pm$0.00 &  \cellcolor[rgb]{0.99,1.00,1.00} 0.00$\pm$0.00 &  \cellcolor[rgb]{0.99,1.00,1.00} 0.00$\pm$0.01 &  \cellcolor[rgb]{0.99,1.00,1.00} 0.00$\pm$0.00 &  \cellcolor[rgb]{0.99,1.00,1.00} 0.00$\pm$0.00 \\
BC (MT)                      &  \cellcolor[rgb]{0.74,0.86,0.99} 0.89$\pm$0.04 &  \cellcolor[rgb]{0.79,0.88,0.97} 0.78$\pm$0.12 &  \cellcolor[rgb]{0.99,0.99,1.00} 0.05$\pm$0.03 &  \cellcolor[rgb]{0.94,0.97,0.98} 0.31$\pm$0.07 &  \cellcolor[rgb]{0.78,0.88,0.97} 0.81$\pm$0.07 &  \cellcolor[rgb]{0.99,0.99,1.00} 0.03$\pm$0.01 &  \cellcolor[rgb]{0.78,0.88,0.97} 0.81$\pm$0.08 &  \cellcolor[rgb]{0.99,0.99,1.00} 0.05$\pm$0.03 \\
BC (ST)                      &  \cellcolor[rgb]{0.74,0.86,0.99} 0.89$\pm$0.03 &  \cellcolor[rgb]{0.80,0.89,0.96} 0.76$\pm$0.02 &  \cellcolor[rgb]{0.99,0.99,1.00} 0.04$\pm$0.01 &  \cellcolor[rgb]{0.86,0.92,0.96} 0.60$\pm$0.04 &  \cellcolor[rgb]{0.78,0.88,0.97} 0.80$\pm$0.02 &  \cellcolor[rgb]{0.98,0.99,1.00} 0.06$\pm$0.02 &  \cellcolor[rgb]{0.79,0.89,0.97} 0.77$\pm$0.04 &  \cellcolor[rgb]{0.99,0.99,1.00} 0.05$\pm$0.01 \\
BC (ST, allo.)               &  \cellcolor[rgb]{0.73,0.85,0.98} 0.93$\pm$0.02 &  \cellcolor[rgb]{0.79,0.88,0.97} 0.79$\pm$0.07 &  \cellcolor[rgb]{0.99,1.00,1.00} 0.01$\pm$0.01 &  \cellcolor[rgb]{0.82,0.90,0.96} 0.72$\pm$0.11 &  \cellcolor[rgb]{0.80,0.89,0.96} 0.75$\pm$0.09 &  \cellcolor[rgb]{0.99,1.00,1.00} 0.02$\pm$0.01 &  \cellcolor[rgb]{0.81,0.89,0.96} 0.74$\pm$0.08 &  \cellcolor[rgb]{0.99,0.99,1.00} 0.02$\pm$0.01 \\
BC (ST, no aug.)             &  \cellcolor[rgb]{0.73,0.84,0.98} 0.94$\pm$0.06 &  \cellcolor[rgb]{0.92,0.96,0.98} 0.38$\pm$0.06 &  \cellcolor[rgb]{0.98,0.99,1.00} 0.06$\pm$0.03 &  \cellcolor[rgb]{0.93,0.96,0.98} 0.36$\pm$0.04 &  \cellcolor[rgb]{0.80,0.89,0.96} 0.75$\pm$0.08 &  \cellcolor[rgb]{0.99,0.99,1.00} 0.04$\pm$0.02 &  \cellcolor[rgb]{0.79,0.88,0.97} 0.78$\pm$0.06 &  \cellcolor[rgb]{0.98,0.99,1.00} 0.06$\pm$0.02 \\
BC (ST, no trans./rot. aug.) &  \cellcolor[rgb]{0.73,0.85,0.98} 0.93$\pm$0.03 &  \cellcolor[rgb]{0.90,0.95,0.97} 0.45$\pm$0.10 &  \cellcolor[rgb]{0.98,0.99,1.00} 0.09$\pm$0.01 &  \cellcolor[rgb]{0.90,0.95,0.97} 0.43$\pm$0.05 &  \cellcolor[rgb]{0.78,0.88,0.97} 0.81$\pm$0.07 &  \cellcolor[rgb]{0.99,0.99,1.00} 0.05$\pm$0.03 &  \cellcolor[rgb]{0.80,0.89,0.96} 0.75$\pm$0.09 &  \cellcolor[rgb]{0.98,0.99,1.00} 0.06$\pm$0.03 \\
GAIL (MT)                    &  \cellcolor[rgb]{0.91,0.95,0.97} 0.43$\pm$0.24 &  \cellcolor[rgb]{0.91,0.95,0.97} 0.41$\pm$0.21 &  \cellcolor[rgb]{0.99,0.99,1.00} 0.02$\pm$0.02 &  \cellcolor[rgb]{0.98,0.99,1.00} 0.06$\pm$0.04 &  \cellcolor[rgb]{0.92,0.96,0.98} 0.38$\pm$0.24 &  \cellcolor[rgb]{0.99,0.99,1.00} 0.03$\pm$0.02 &  \cellcolor[rgb]{0.92,0.96,0.98} 0.39$\pm$0.24 &  \cellcolor[rgb]{0.99,0.99,1.00} 0.02$\pm$0.02 \\
GAIL (ST)                    &  \cellcolor[rgb]{0.72,0.83,0.98} 0.98$\pm$0.02 &  \cellcolor[rgb]{0.72,0.83,0.98} 0.97$\pm$0.01 &  \cellcolor[rgb]{0.98,0.99,1.00} 0.10$\pm$0.02 &  \cellcolor[rgb]{0.96,0.97,0.99} 0.23$\pm$0.06 &  \cellcolor[rgb]{0.72,0.84,0.98} 0.95$\pm$0.04 &  \cellcolor[rgb]{0.99,0.99,1.00} 0.05$\pm$0.02 &  \cellcolor[rgb]{0.72,0.84,0.98} 0.96$\pm$0.02 &  \cellcolor[rgb]{0.99,0.99,1.00} 0.05$\pm$0.01 \\
GAIL (ST, allo.)             &  \cellcolor[rgb]{0.72,0.84,0.98} 0.95$\pm$0.03 &  \cellcolor[rgb]{0.76,0.87,0.98} 0.84$\pm$0.03 &  \cellcolor[rgb]{0.99,1.00,1.00} 0.00$\pm$0.00 &  \cellcolor[rgb]{0.88,0.94,0.97} 0.50$\pm$0.09 &  \cellcolor[rgb]{0.74,0.86,0.99} 0.87$\pm$0.05 &  \cellcolor[rgb]{0.99,1.00,1.00} 0.00$\pm$0.01 &  \cellcolor[rgb]{0.73,0.86,0.99} 0.90$\pm$0.04 &  \cellcolor[rgb]{0.99,1.00,1.00} 0.01$\pm$0.01 \\
GAIL (ST, no aug.)           &  \cellcolor[rgb]{0.90,0.94,0.97} 0.46$\pm$0.28 &  \cellcolor[rgb]{0.93,0.96,0.98} 0.36$\pm$0.27 &  \cellcolor[rgb]{0.99,1.00,1.00} 0.01$\pm$0.01 &  \cellcolor[rgb]{0.98,0.99,1.00} 0.10$\pm$0.07 &  \cellcolor[rgb]{0.93,0.96,0.98} 0.34$\pm$0.25 &  \cellcolor[rgb]{0.99,0.99,1.00} 0.02$\pm$0.01 &  \cellcolor[rgb]{0.93,0.96,0.98} 0.37$\pm$0.27 &  \cellcolor[rgb]{0.99,1.00,1.00} 0.02$\pm$0.01 \\
WGAIL-GP (ST)                &  \cellcolor[rgb]{0.83,0.90,0.96} 0.70$\pm$0.09 &  \cellcolor[rgb]{0.90,0.94,0.97} 0.46$\pm$0.11 &  \cellcolor[rgb]{0.99,0.99,1.00} 0.02$\pm$0.00 &  \cellcolor[rgb]{0.98,0.99,1.00} 0.09$\pm$0.04 &  \cellcolor[rgb]{0.85,0.91,0.95} 0.64$\pm$0.04 &  \cellcolor[rgb]{0.99,1.00,1.00} 0.02$\pm$0.01 &  \cellcolor[rgb]{0.85,0.91,0.96} 0.62$\pm$0.03 &  \cellcolor[rgb]{0.99,0.99,1.00} 0.03$\pm$0.02 \\
\midrule
     & \multicolumn{8}{c}{\textbf{FixColour}} \\
        &                                           Demo &                                         Jitter &                                         Layout &                                         Colour &                                          Shape &                                      CountPlus &                                       Dynamics &                                            All \\
\midrule
AL (ST)                      &  \cellcolor[rgb]{0.99,1.00,1.00} 0.00$\pm$0.01 &  \cellcolor[rgb]{0.99,0.99,1.00} 0.03$\pm$0.04 &  \cellcolor[rgb]{0.99,0.99,1.00} 0.02$\pm$0.01 &  \cellcolor[rgb]{0.99,0.99,1.00} 0.03$\pm$0.03 &  \cellcolor[rgb]{0.99,1.00,1.00} 0.00$\pm$0.01 &  \cellcolor[rgb]{0.99,1.00,1.00} 0.01$\pm$0.01 &  \cellcolor[rgb]{0.99,1.00,1.00} 0.01$\pm$0.01 &  \cellcolor[rgb]{0.99,1.00,1.00} 0.01$\pm$0.02 \\
BC (MT)                      &  \cellcolor[rgb]{0.80,0.89,0.97} 0.76$\pm$0.18 &  \cellcolor[rgb]{0.86,0.92,0.96} 0.61$\pm$0.14 &  \cellcolor[rgb]{0.97,0.98,0.99} 0.15$\pm$0.03 &  \cellcolor[rgb]{0.96,0.97,0.99} 0.23$\pm$0.05 &  \cellcolor[rgb]{0.82,0.90,0.96} 0.72$\pm$0.18 &  \cellcolor[rgb]{0.97,0.98,0.99} 0.13$\pm$0.02 &  \cellcolor[rgb]{0.83,0.90,0.96} 0.70$\pm$0.21 &  \cellcolor[rgb]{0.97,0.98,0.99} 0.14$\pm$0.07 \\
BC (ST)                      &  \cellcolor[rgb]{0.85,0.91,0.96} 0.62$\pm$0.14 &  \cellcolor[rgb]{0.90,0.95,0.97} 0.45$\pm$0.18 &  \cellcolor[rgb]{0.97,0.98,0.99} 0.18$\pm$0.04 &  \cellcolor[rgb]{0.96,0.98,0.99} 0.19$\pm$0.03 &  \cellcolor[rgb]{0.85,0.91,0.96} 0.63$\pm$0.12 &  \cellcolor[rgb]{0.97,0.98,0.99} 0.18$\pm$0.04 &  \cellcolor[rgb]{0.82,0.90,0.96} 0.71$\pm$0.13 &  \cellcolor[rgb]{0.97,0.98,0.99} 0.13$\pm$0.03 \\
BC (ST, allo.)               &  \cellcolor[rgb]{0.74,0.86,0.99} 0.88$\pm$0.07 &  \cellcolor[rgb]{0.89,0.94,0.97} 0.47$\pm$0.24 &  \cellcolor[rgb]{0.98,0.99,0.99} 0.11$\pm$0.03 &  \cellcolor[rgb]{0.94,0.96,0.98} 0.32$\pm$0.03 &  \cellcolor[rgb]{0.75,0.87,0.98} 0.86$\pm$0.06 &  \cellcolor[rgb]{0.97,0.98,0.99} 0.13$\pm$0.04 &  \cellcolor[rgb]{0.76,0.87,0.98} 0.84$\pm$0.04 &  \cellcolor[rgb]{0.97,0.98,0.99} 0.14$\pm$0.05 \\
BC (ST, no aug.)             &  \cellcolor[rgb]{0.87,0.93,0.96} 0.55$\pm$0.18 &  \cellcolor[rgb]{0.95,0.97,0.98} 0.29$\pm$0.12 &  \cellcolor[rgb]{0.96,0.98,0.99} 0.19$\pm$0.03 &  \cellcolor[rgb]{0.96,0.98,0.99} 0.21$\pm$0.09 &  \cellcolor[rgb]{0.87,0.92,0.96} 0.57$\pm$0.18 &  \cellcolor[rgb]{0.96,0.97,0.99} 0.22$\pm$0.03 &  \cellcolor[rgb]{0.88,0.93,0.97} 0.52$\pm$0.23 &  \cellcolor[rgb]{0.96,0.98,0.99} 0.22$\pm$0.02 \\
BC (ST, no trans./rot. aug.) &  \cellcolor[rgb]{0.90,0.94,0.97} 0.46$\pm$0.14 &  \cellcolor[rgb]{0.95,0.97,0.98} 0.28$\pm$0.07 &  \cellcolor[rgb]{0.97,0.98,0.99} 0.17$\pm$0.04 &  \cellcolor[rgb]{0.97,0.98,0.99} 0.17$\pm$0.07 &  \cellcolor[rgb]{0.89,0.94,0.97} 0.48$\pm$0.16 &  \cellcolor[rgb]{0.96,0.97,0.99} 0.23$\pm$0.03 &  \cellcolor[rgb]{0.92,0.96,0.98} 0.39$\pm$0.13 &  \cellcolor[rgb]{0.96,0.98,0.99} 0.20$\pm$0.04 \\
GAIL (MT)                    &  \cellcolor[rgb]{0.71,0.82,0.98} 0.99$\pm$0.02 &  \cellcolor[rgb]{0.85,0.91,0.95} 0.65$\pm$0.17 &  \cellcolor[rgb]{0.94,0.97,0.98} 0.30$\pm$0.08 &  \cellcolor[rgb]{0.96,0.98,0.99} 0.21$\pm$0.06 &  \cellcolor[rgb]{0.71,0.82,0.98} 0.99$\pm$0.01 &  \cellcolor[rgb]{0.97,0.98,0.99} 0.18$\pm$0.10 &  \cellcolor[rgb]{0.72,0.84,0.98} 0.95$\pm$0.07 &  \cellcolor[rgb]{0.97,0.98,0.99} 0.16$\pm$0.06 \\
GAIL (ST)                    &  \cellcolor[rgb]{0.71,0.82,0.98} 0.99$\pm$0.01 &  \cellcolor[rgb]{0.76,0.87,0.98} 0.84$\pm$0.07 &  \cellcolor[rgb]{0.94,0.96,0.98} 0.32$\pm$0.07 &  \cellcolor[rgb]{0.96,0.97,0.99} 0.25$\pm$0.04 &  \cellcolor[rgb]{0.72,0.83,0.98} 0.97$\pm$0.03 &  \cellcolor[rgb]{0.96,0.98,0.99} 0.19$\pm$0.02 &  \cellcolor[rgb]{0.72,0.83,0.98} 0.96$\pm$0.02 &  \cellcolor[rgb]{0.96,0.98,0.99} 0.20$\pm$0.01 \\
GAIL (ST, allo.)             &  \cellcolor[rgb]{0.72,0.83,0.98} 0.98$\pm$0.00 &  \cellcolor[rgb]{0.84,0.91,0.95} 0.66$\pm$0.21 &  \cellcolor[rgb]{0.99,0.99,1.00} 0.06$\pm$0.02 &  \cellcolor[rgb]{0.93,0.96,0.98} 0.36$\pm$0.01 &  \cellcolor[rgb]{0.71,0.82,0.98} 0.99$\pm$0.01 &  \cellcolor[rgb]{0.98,0.99,1.00} 0.09$\pm$0.04 &  \cellcolor[rgb]{0.72,0.83,0.98} 0.98$\pm$0.01 &  \cellcolor[rgb]{0.98,0.99,1.00} 0.09$\pm$0.03 \\
GAIL (ST, no aug.)           &  \cellcolor[rgb]{0.72,0.83,0.98} 0.98$\pm$0.02 &  \cellcolor[rgb]{0.85,0.91,0.95} 0.64$\pm$0.12 &  \cellcolor[rgb]{0.97,0.98,0.99} 0.16$\pm$0.05 &  \cellcolor[rgb]{0.95,0.97,0.99} 0.27$\pm$0.03 &  \cellcolor[rgb]{0.74,0.86,0.99} 0.87$\pm$0.07 &  \cellcolor[rgb]{0.97,0.98,0.99} 0.13$\pm$0.04 &  \cellcolor[rgb]{0.76,0.87,0.98} 0.84$\pm$0.07 &  \cellcolor[rgb]{0.97,0.98,0.99} 0.14$\pm$0.03 \\
WGAIL-GP (ST)                &  \cellcolor[rgb]{0.73,0.85,0.98} 0.93$\pm$0.04 &  \cellcolor[rgb]{0.82,0.90,0.96} 0.72$\pm$0.12 &  \cellcolor[rgb]{0.98,0.99,1.00} 0.08$\pm$0.03 &  \cellcolor[rgb]{0.97,0.98,0.99} 0.18$\pm$0.02 &  \cellcolor[rgb]{0.73,0.85,0.98} 0.93$\pm$0.04 &  \cellcolor[rgb]{0.99,1.00,1.00} 0.02$\pm$0.02 &  \cellcolor[rgb]{0.73,0.85,0.99} 0.91$\pm$0.05 &  \cellcolor[rgb]{0.99,0.99,1.00} 0.05$\pm$0.02 \\
\bottomrule
\end{tabular}
        \end{footnotesize}
        \vspace{3mm}
        \caption{
            Additional results; refer to \cref{tab:results-tab1} for details.
        }\label{tab:results-tab3}
    \end{table}
\end{landscape}

\begin{landscape}
    \begin{table}
        \centering
        \begin{footnotesize}
            \begin{tabular}{lcccccccc}
\toprule
     & \multicolumn{8}{c}{\textbf{ClusterColour}} \\
        &                                           Demo &                                         Jitter &                                         Layout &                                         Colour &                                          Shape &                                      CountPlus &                                       Dynamics &                                            All \\
\midrule
AL (ST)                      &  \cellcolor[rgb]{0.99,1.00,1.00} 0.00$\pm$0.00 &  \cellcolor[rgb]{0.99,1.00,1.00} 0.00$\pm$0.00 &  \cellcolor[rgb]{0.99,1.00,1.00} 0.00$\pm$0.00 &  \cellcolor[rgb]{0.99,1.00,1.00} 0.01$\pm$0.01 &  \cellcolor[rgb]{0.99,1.00,1.00} 0.00$\pm$0.00 &  \cellcolor[rgb]{0.99,1.00,1.00} 0.00$\pm$0.00 &  \cellcolor[rgb]{0.99,1.00,1.00} 0.00$\pm$0.00 &  \cellcolor[rgb]{0.99,1.00,1.00} 0.00$\pm$0.00 \\
BC (MT)                      &  \cellcolor[rgb]{0.98,0.99,0.99} 0.11$\pm$0.05 &  \cellcolor[rgb]{0.98,0.99,1.00} 0.10$\pm$0.03 &  \cellcolor[rgb]{0.99,1.00,1.00} 0.00$\pm$0.01 &  \cellcolor[rgb]{0.99,1.00,1.00} 0.01$\pm$0.01 &  \cellcolor[rgb]{0.98,0.99,1.00} 0.10$\pm$0.05 &  \cellcolor[rgb]{0.99,1.00,1.00} 0.00$\pm$0.00 &  \cellcolor[rgb]{0.98,0.99,1.00} 0.09$\pm$0.06 &  \cellcolor[rgb]{0.99,1.00,1.00} 0.01$\pm$0.00 \\
BC (ST)                      &  \cellcolor[rgb]{0.97,0.98,0.99} 0.16$\pm$0.05 &  \cellcolor[rgb]{0.97,0.98,0.99} 0.17$\pm$0.04 &  \cellcolor[rgb]{0.99,1.00,1.00} 0.01$\pm$0.01 &  \cellcolor[rgb]{0.99,1.00,1.00} 0.01$\pm$0.00 &  \cellcolor[rgb]{0.97,0.98,0.99} 0.17$\pm$0.04 &  \cellcolor[rgb]{0.99,1.00,1.00} 0.01$\pm$0.01 &  \cellcolor[rgb]{0.97,0.98,0.99} 0.16$\pm$0.05 &  \cellcolor[rgb]{0.99,1.00,1.00} 0.01$\pm$0.00 \\
BC (ST, allo.)               &  \cellcolor[rgb]{0.98,0.99,1.00} 0.10$\pm$0.05 &  \cellcolor[rgb]{0.98,0.99,0.99} 0.11$\pm$0.02 &  \cellcolor[rgb]{0.99,1.00,1.00} 0.00$\pm$0.00 &  \cellcolor[rgb]{0.99,1.00,1.00} 0.00$\pm$0.00 &  \cellcolor[rgb]{0.98,0.99,1.00} 0.10$\pm$0.04 &  \cellcolor[rgb]{0.99,1.00,1.00} 0.00$\pm$0.00 &  \cellcolor[rgb]{0.98,0.99,1.00} 0.07$\pm$0.03 &  \cellcolor[rgb]{0.99,1.00,1.00} 0.00$\pm$0.00 \\
BC (ST, no aug.)             &  \cellcolor[rgb]{0.99,0.99,1.00} 0.04$\pm$0.02 &  \cellcolor[rgb]{0.99,1.00,1.00} 0.02$\pm$0.01 &  \cellcolor[rgb]{0.99,1.00,1.00} 0.01$\pm$0.01 &  \cellcolor[rgb]{0.99,1.00,1.00} 0.01$\pm$0.01 &  \cellcolor[rgb]{0.99,1.00,1.00} 0.02$\pm$0.01 &  \cellcolor[rgb]{0.99,1.00,1.00} 0.00$\pm$0.00 &  \cellcolor[rgb]{0.99,0.99,1.00} 0.04$\pm$0.01 &  \cellcolor[rgb]{0.99,1.00,1.00} 0.00$\pm$0.00 \\
BC (ST, no trans./rot. aug.) &  \cellcolor[rgb]{0.99,0.99,1.00} 0.05$\pm$0.01 &  \cellcolor[rgb]{0.99,1.00,1.00} 0.02$\pm$0.00 &  \cellcolor[rgb]{0.99,1.00,1.00} 0.00$\pm$0.00 &  \cellcolor[rgb]{0.99,1.00,1.00} 0.01$\pm$0.01 &  \cellcolor[rgb]{0.99,0.99,1.00} 0.04$\pm$0.02 &  \cellcolor[rgb]{0.99,1.00,1.00} 0.00$\pm$0.00 &  \cellcolor[rgb]{0.99,0.99,1.00} 0.04$\pm$0.02 &  \cellcolor[rgb]{0.99,1.00,1.00} 0.00$\pm$0.00 \\
GAIL (MT)                    &  \cellcolor[rgb]{0.99,1.00,1.00} 0.01$\pm$0.01 &  \cellcolor[rgb]{0.99,1.00,1.00} 0.01$\pm$0.01 &  \cellcolor[rgb]{0.99,1.00,1.00} 0.01$\pm$0.00 &  \cellcolor[rgb]{0.99,1.00,1.00} 0.01$\pm$0.00 &  \cellcolor[rgb]{0.99,1.00,1.00} 0.01$\pm$0.01 &  \cellcolor[rgb]{0.99,1.00,1.00} 0.00$\pm$0.00 &  \cellcolor[rgb]{0.99,1.00,1.00} 0.01$\pm$0.01 &  \cellcolor[rgb]{0.99,1.00,1.00} 0.00$\pm$0.00 \\
GAIL (ST)                    &  \cellcolor[rgb]{0.98,0.99,0.99} 0.12$\pm$0.03 &  \cellcolor[rgb]{0.98,0.99,0.99} 0.11$\pm$0.04 &  \cellcolor[rgb]{0.99,1.00,1.00} 0.01$\pm$0.00 &  \cellcolor[rgb]{0.99,1.00,1.00} 0.01$\pm$0.00 &  \cellcolor[rgb]{0.98,0.99,1.00} 0.08$\pm$0.04 &  \cellcolor[rgb]{0.99,1.00,1.00} 0.01$\pm$0.00 &  \cellcolor[rgb]{0.98,0.99,0.99} 0.11$\pm$0.03 &  \cellcolor[rgb]{0.99,1.00,1.00} 0.01$\pm$0.01 \\
GAIL (ST, allo.)             &  \cellcolor[rgb]{0.99,1.00,1.00} 0.00$\pm$0.00 &  \cellcolor[rgb]{0.99,1.00,1.00} 0.00$\pm$0.00 &  \cellcolor[rgb]{0.99,1.00,1.00} 0.00$\pm$0.00 &  \cellcolor[rgb]{0.99,1.00,1.00} 0.00$\pm$0.00 &  \cellcolor[rgb]{0.99,1.00,1.00} 0.01$\pm$0.01 &  \cellcolor[rgb]{0.99,1.00,1.00} 0.00$\pm$0.00 &  \cellcolor[rgb]{0.99,1.00,1.00} 0.01$\pm$0.01 &  \cellcolor[rgb]{0.99,1.00,1.00} 0.00$\pm$0.00 \\
GAIL (ST, no aug.)           &  \cellcolor[rgb]{0.99,0.99,1.00} 0.02$\pm$0.01 &  \cellcolor[rgb]{0.99,1.00,1.00} 0.01$\pm$0.01 &  \cellcolor[rgb]{0.99,1.00,1.00} 0.00$\pm$0.00 &  \cellcolor[rgb]{0.99,1.00,1.00} 0.01$\pm$0.00 &  \cellcolor[rgb]{0.99,1.00,1.00} 0.02$\pm$0.01 &  \cellcolor[rgb]{0.99,1.00,1.00} 0.01$\pm$0.01 &  \cellcolor[rgb]{0.99,0.99,1.00} 0.02$\pm$0.01 &  \cellcolor[rgb]{0.99,1.00,1.00} 0.00$\pm$0.00 \\
WGAIL-GP (ST)                &  \cellcolor[rgb]{0.99,1.00,1.00} 0.00$\pm$0.00 &  \cellcolor[rgb]{0.99,1.00,1.00} 0.00$\pm$0.00 &  \cellcolor[rgb]{0.99,1.00,1.00} 0.00$\pm$0.00 &  \cellcolor[rgb]{0.99,1.00,1.00} 0.01$\pm$0.01 &  \cellcolor[rgb]{0.99,1.00,1.00} 0.00$\pm$0.00 &  \cellcolor[rgb]{0.99,1.00,1.00} 0.00$\pm$0.00 &  \cellcolor[rgb]{0.99,1.00,1.00} 0.00$\pm$0.00 &  \cellcolor[rgb]{0.99,1.00,1.00} 0.00$\pm$0.00 \\
\midrule
     & \multicolumn{8}{c}{\textbf{ClusterShape}} \\
        &                                           Demo &                                         Jitter &                                         Layout &                                         Colour &                                          Shape &                                      CountPlus &                                       Dynamics &                                            All \\
\midrule
AL (ST)                      &  \cellcolor[rgb]{0.99,1.00,1.00} 0.00$\pm$0.00 &  \cellcolor[rgb]{0.99,1.00,1.00} 0.00$\pm$0.00 &  \cellcolor[rgb]{0.99,1.00,1.00} 0.00$\pm$0.00 &  \cellcolor[rgb]{0.99,1.00,1.00} 0.00$\pm$0.00 &  \cellcolor[rgb]{0.99,1.00,1.00} 0.00$\pm$0.00 &  \cellcolor[rgb]{0.99,1.00,1.00} 0.00$\pm$0.00 &  \cellcolor[rgb]{0.99,1.00,1.00} 0.00$\pm$0.00 &  \cellcolor[rgb]{0.99,1.00,1.00} 0.01$\pm$0.00 \\
BC (MT)                      &  \cellcolor[rgb]{0.96,0.97,0.99} 0.23$\pm$0.06 &  \cellcolor[rgb]{0.96,0.98,0.99} 0.19$\pm$0.05 &  \cellcolor[rgb]{0.99,1.00,1.00} 0.00$\pm$0.00 &  \cellcolor[rgb]{0.99,0.99,1.00} 0.02$\pm$0.01 &  \cellcolor[rgb]{0.99,1.00,1.00} 0.00$\pm$0.00 &  \cellcolor[rgb]{0.99,1.00,1.00} 0.00$\pm$0.00 &  \cellcolor[rgb]{0.96,0.98,0.99} 0.19$\pm$0.04 &  \cellcolor[rgb]{0.99,1.00,1.00} 0.01$\pm$0.00 \\
BC (ST)                      &  \cellcolor[rgb]{0.93,0.96,0.98} 0.35$\pm$0.11 &  \cellcolor[rgb]{0.95,0.97,0.99} 0.26$\pm$0.06 &  \cellcolor[rgb]{0.99,1.00,1.00} 0.01$\pm$0.01 &  \cellcolor[rgb]{0.98,0.99,1.00} 0.06$\pm$0.02 &  \cellcolor[rgb]{0.99,1.00,1.00} 0.01$\pm$0.00 &  \cellcolor[rgb]{0.99,1.00,1.00} 0.00$\pm$0.00 &  \cellcolor[rgb]{0.94,0.97,0.98} 0.31$\pm$0.04 &  \cellcolor[rgb]{0.99,1.00,1.00} 0.01$\pm$0.01 \\
BC (ST, allo.)               &  \cellcolor[rgb]{0.96,0.98,0.99} 0.21$\pm$0.02 &  \cellcolor[rgb]{0.96,0.97,0.99} 0.23$\pm$0.10 &  \cellcolor[rgb]{0.99,1.00,1.00} 0.00$\pm$0.00 &  \cellcolor[rgb]{0.97,0.98,0.99} 0.13$\pm$0.04 &  \cellcolor[rgb]{0.99,1.00,1.00} 0.00$\pm$0.00 &  \cellcolor[rgb]{0.99,1.00,1.00} 0.00$\pm$0.00 &  \cellcolor[rgb]{0.97,0.98,0.99} 0.16$\pm$0.05 &  \cellcolor[rgb]{0.99,1.00,1.00} 0.01$\pm$0.00 \\
BC (ST, no aug.)             &  \cellcolor[rgb]{0.98,0.99,0.99} 0.12$\pm$0.03 &  \cellcolor[rgb]{0.98,0.99,1.00} 0.10$\pm$0.02 &  \cellcolor[rgb]{0.99,1.00,1.00} 0.00$\pm$0.01 &  \cellcolor[rgb]{0.99,1.00,1.00} 0.02$\pm$0.01 &  \cellcolor[rgb]{0.99,1.00,1.00} 0.01$\pm$0.01 &  \cellcolor[rgb]{0.99,1.00,1.00} 0.00$\pm$0.01 &  \cellcolor[rgb]{0.98,0.99,1.00} 0.09$\pm$0.04 &  \cellcolor[rgb]{0.99,1.00,1.00} 0.01$\pm$0.01 \\
BC (ST, no trans./rot. aug.) &  \cellcolor[rgb]{0.98,0.99,0.99} 0.11$\pm$0.05 &  \cellcolor[rgb]{0.98,0.99,0.99} 0.11$\pm$0.02 &  \cellcolor[rgb]{0.99,1.00,1.00} 0.00$\pm$0.00 &  \cellcolor[rgb]{0.99,1.00,1.00} 0.01$\pm$0.01 &  \cellcolor[rgb]{0.99,1.00,1.00} 0.00$\pm$0.00 &  \cellcolor[rgb]{0.99,1.00,1.00} 0.00$\pm$0.01 &  \cellcolor[rgb]{0.98,0.99,0.99} 0.11$\pm$0.04 &  \cellcolor[rgb]{0.99,1.00,1.00} 0.01$\pm$0.00 \\
GAIL (MT)                    &  \cellcolor[rgb]{0.99,1.00,1.00} 0.01$\pm$0.01 &  \cellcolor[rgb]{0.99,1.00,1.00} 0.01$\pm$0.01 &  \cellcolor[rgb]{0.99,1.00,1.00} 0.01$\pm$0.00 &  \cellcolor[rgb]{0.99,1.00,1.00} 0.00$\pm$0.00 &  \cellcolor[rgb]{0.99,1.00,1.00} 0.00$\pm$0.00 &  \cellcolor[rgb]{0.99,1.00,1.00} 0.00$\pm$0.00 &  \cellcolor[rgb]{0.99,1.00,1.00} 0.01$\pm$0.02 &  \cellcolor[rgb]{0.99,1.00,1.00} 0.01$\pm$0.01 \\
GAIL (ST)                    &  \cellcolor[rgb]{0.90,0.95,0.97} 0.44$\pm$0.05 &  \cellcolor[rgb]{0.90,0.95,0.97} 0.44$\pm$0.08 &  \cellcolor[rgb]{0.99,1.00,1.00} 0.01$\pm$0.02 &  \cellcolor[rgb]{0.99,1.00,1.00} 0.02$\pm$0.01 &  \cellcolor[rgb]{0.99,1.00,1.00} 0.01$\pm$0.00 &  \cellcolor[rgb]{0.99,1.00,1.00} 0.00$\pm$0.00 &  \cellcolor[rgb]{0.92,0.96,0.98} 0.39$\pm$0.01 &  \cellcolor[rgb]{0.99,1.00,1.00} 0.01$\pm$0.00 \\
GAIL (ST, allo.)             &  \cellcolor[rgb]{0.99,1.00,1.00} 0.00$\pm$0.00 &  \cellcolor[rgb]{0.99,1.00,1.00} 0.00$\pm$0.00 &  \cellcolor[rgb]{0.99,1.00,1.00} 0.01$\pm$0.01 &  \cellcolor[rgb]{0.99,1.00,1.00} 0.00$\pm$0.00 &  \cellcolor[rgb]{0.99,1.00,1.00} 0.00$\pm$0.00 &  \cellcolor[rgb]{0.99,1.00,1.00} 0.00$\pm$0.00 &  \cellcolor[rgb]{0.99,1.00,1.00} 0.00$\pm$0.00 &  \cellcolor[rgb]{0.99,1.00,1.00} 0.01$\pm$0.00 \\
GAIL (ST, no aug.)           &  \cellcolor[rgb]{0.99,1.00,1.00} 0.01$\pm$0.02 &  \cellcolor[rgb]{0.99,1.00,1.00} 0.02$\pm$0.01 &  \cellcolor[rgb]{0.99,1.00,1.00} 0.00$\pm$0.00 &  \cellcolor[rgb]{0.99,1.00,1.00} 0.00$\pm$0.00 &  \cellcolor[rgb]{0.99,1.00,1.00} 0.00$\pm$0.01 &  \cellcolor[rgb]{0.99,1.00,1.00} 0.01$\pm$0.00 &  \cellcolor[rgb]{0.99,1.00,1.00} 0.01$\pm$0.01 &  \cellcolor[rgb]{0.99,1.00,1.00} 0.01$\pm$0.00 \\
WGAIL-GP (ST)                &  \cellcolor[rgb]{0.99,1.00,1.00} 0.01$\pm$0.00 &  \cellcolor[rgb]{0.99,1.00,1.00} 0.01$\pm$0.00 &  \cellcolor[rgb]{0.99,1.00,1.00} 0.01$\pm$0.01 &  \cellcolor[rgb]{0.99,1.00,1.00} 0.00$\pm$0.00 &  \cellcolor[rgb]{0.99,1.00,1.00} 0.00$\pm$0.00 &  \cellcolor[rgb]{0.99,1.00,1.00} 0.00$\pm$0.00 &  \cellcolor[rgb]{0.99,1.00,1.00} 0.00$\pm$0.00 &  \cellcolor[rgb]{0.99,1.00,1.00} 0.00$\pm$0.00 \\
\bottomrule
\end{tabular}
        \end{footnotesize}
        \vspace{3mm}
        \caption{
            Additional results; refer to \cref{tab:results-tab1} for details.
        }\label{tab:results-tab4}
    \end{table}
\end{landscape}

\end{document}